\documentclass[11pt]{article}

\usepackage[final]{acl}

\usepackage{times}
\usepackage{latexsym}

\usepackage{hyperref}
\usepackage{url}

\usepackage{pifont}
\usepackage{xspace}
\usepackage{enumitem}
\usepackage{amsthm}
\usepackage{graphicx}
\usepackage{amssymb}
\usepackage{multirow}
\usepackage{booktabs}
\usepackage{scalerel}
\usepackage{makecell}
\usepackage{wrapfig}
\usepackage{colortbl}
\usepackage{xcolor} % For color support
\usepackage{soul} % For highlighting
\usepackage[normalem]{ulem}
\usepackage{microtype} 
\usepackage{caption}
\usepackage{makecell}
\usepackage{array}
\usepackage{booktabs}
\usepackage[table]{xcolor}
\usepackage{graphicx}
\usepackage{booktabs,tabularx,array}
\newcolumntype{L}[1]{>{\raggedright\arraybackslash}p{#1}} % 可选：手动宽度的左对齐列
\usepackage{bbm}
\usepackage{booktabs}
\usepackage{multirow}
\usepackage{amssymb}    % 提供 \checkmark
\usepackage{graphicx}   % 提供 \scalebox
\usepackage{siunitx}
\usepackage{amsmath}

\usepackage[T1]{fontenc}
\usepackage[utf8]{inputenc}

\usepackage{microtype}

\usepackage{inconsolata}

\usepackage{graphicx}
\usepackage{booktabs}
\usepackage{siunitx}
\usepackage{enumitem}    % For customizing lists
\usepackage[most]{tcolorbox} % The key package for creating the box

\title{MMTutorBench: The First Multimodal Benchmark for AI Math Tutoring}

\author{
 \textbf{Tengchao Yang\textsuperscript{1}}\thanks{Equal contribution.},
 \textbf{Sichen Guo\textsuperscript{4}}\footnotemark[1],
 \textbf{Mengzhao Jia\textsuperscript{3}},
\\
 \textbf{Jiaming Su\textsuperscript{2}},
 \textbf{Yuanyang Liu\textsuperscript{4}},
 \textbf{Zhihan Zhang\textsuperscript{3}},
 \textbf{Meng Jiang\textsuperscript{3}}\thanks{Corresponding author.}
\\
 \textsuperscript{1}Tongji University
 \textsuperscript{2}Fudan University
 \textsuperscript{3}University of Notre Dame
\\
 \textsuperscript{4}Nanjing University of Posts and Telecommunications
\\
 \texttt{2151298@tongji.edu.cn},
 \texttt{q22010218@njupt.edu.cn}, 
\\
 \texttt{\{mjia2, zzhang23, mjiang2\}@nd.edu} \\
}

\begin{document}

\maketitle
\vspace{-1cm}
\begin{abstract}
Effective math tutoring requires not only solving problems but also diagnosing students' difficulties and guiding them step by step. While multimodal large language models (MLLMs) show promise, existing benchmarks largely overlook these tutoring skills. We introduce MMTutorBench, the first benchmark for AI math tutoring, consisting of 770 problems built around pedagogically significant key-steps. Each problem is paired with problem-specific rubrics that enable fine-grained evaluation across six dimensions, and structured into three tasks—Insight Discovery, Operation Formulation, and Operation Execution. We evaluate 12 leading MLLMs and find clear performance gaps between proprietary and open-source systems, substantial room compared to human tutors, and consistent trends across input variants: OCR pipelines degrade tutoring quality, few-shot prompting yields limited gains, and our rubric-based LLM-as-a-Judge proves highly reliable. These results highlight both the difficulty and diagnostic value of MMTutorBench for advancing AI tutoring. 
Our code and data are available at \url{https://github.com/TangciuYueng/MMTutorBench}.
\end{abstract}

\section{Introduction}

\begin{figure*}[t]
    \centering
    \includegraphics[width=0.95\textwidth]{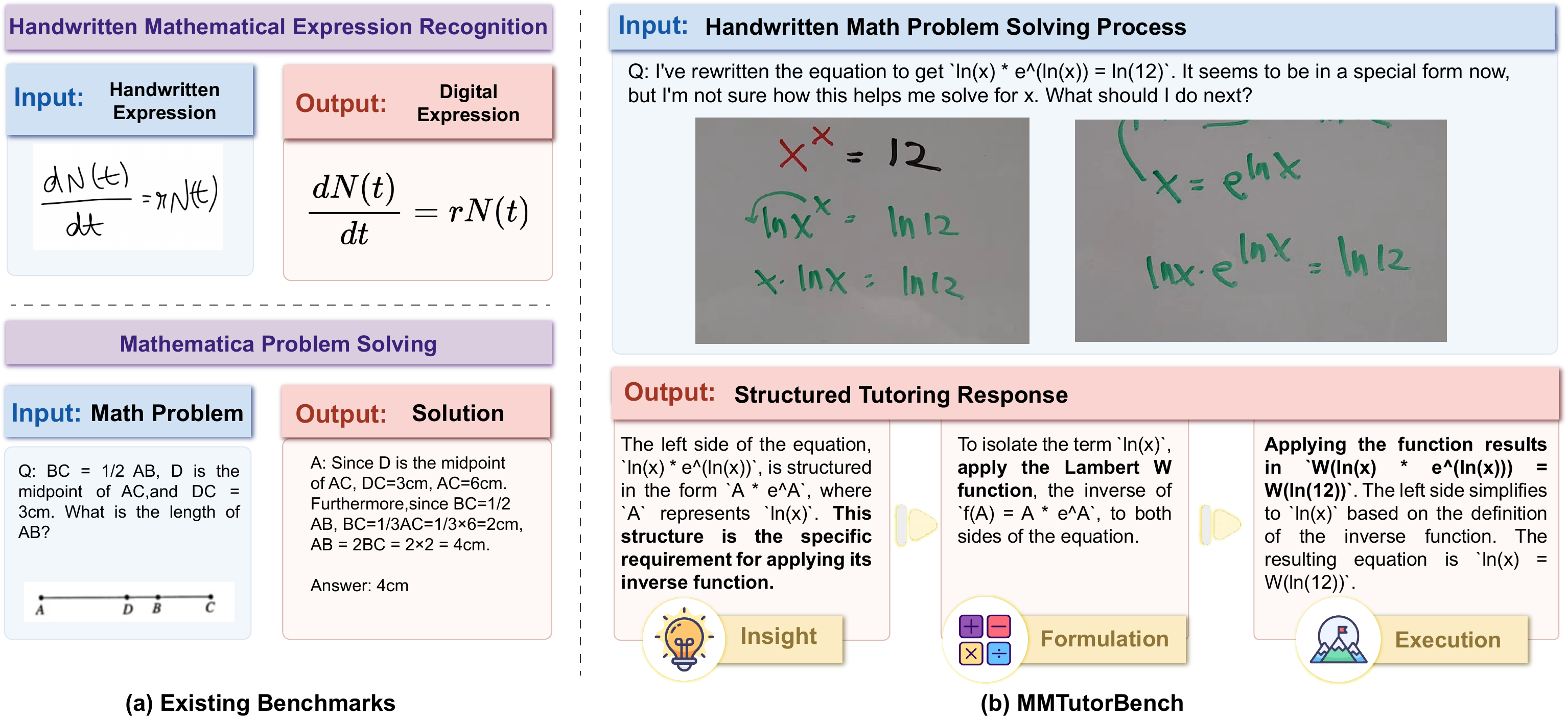}
    \caption{\textbf{(a)} Existing benchmarks usually target a single perspective, such as handwritten expression recognition or problem solving, which is insufficient for evaluating tutoring ability in real educational settings. \textbf{(b)} An example from MMTutorBench: we model the tutoring process in realistic classroom scenarios by taking a student’s handwritten solution attempt and help-seeking question as input. The tutoring response is structured along three dimensions: Insight, Formulation, and Execution. We emphasize some key guidance in bold for illustration. }
    \vspace{-0.5cm}
    \label{fig:intro} 
\end{figure*}

Math tutoring is one of the most important pillars of K-12 education. Many children either lack proper guidance in learning mathematics or receive ineffective support. Psychology research shows that such experiences can lead to ``math anxiety.'' In mild cases, this anxiety causes children to lose confidence in learning math; in more severe cases, it dampens their motivation to learn knowledge and skills more broadly~\cite{wigfield1988math, ashcraft2002math, barroso2021meta}. Studies further reveal that parents themselves often experience anxiety when helping their children with math, and math-anxious parents can unintentionally undermine their children's performance, which can create an unconstructive cycle for children's math learning~\cite{oh2022parents}.

Can AI assist with math tutoring? Being an effective math tutor is non-trivial. Imagine a child working on a problem but getting stuck or making mistakes. For a human or an intelligent system to help effectively, it must have at least five key abilities. First, it needs to ``see'' the problem clearly, recognizing what is being asked. Second, it must understand the problem, apply knowledge, and use chain-of-thought reasoning to solve it correctly. Third, and more importantly, it should interpret \emph{why} the child is struggling by analyzing the context of the child's problem-solving process and identifying the core concepts or ideas that need clarification. Fourth, it should then clarify \emph{what} mathematical operation or method connects to that concept or idea. Finally, it should provide guidance on \emph{how} to take the next concrete step, enabling the child to continue independently, rather than simply revealing the full solution.

Apparently, such an intelligent system would need to be a multimodal large language model (MLLM), and benchmarking is the primary way to evaluate its abilities. For the first two abilities, there already exist related benchmarks. For example, HME100K~\cite{yuan2022syntax}, OCRBench~\cite{fu2024ocrbench}, and MathWriting~\cite{gervais2025mathwriting} can evaluate an MLLM's capacity to extract handwritten text (including math formulas) from images. Math-Vision~\cite{wang2024measuring}, MM-Math~\cite{sun2024mm}, and others can assess a model's ability to solve math problems at various levels. However, when we focus on the three core tasks of AI math tutoring, namely, identifying the key insights, key operations, and next steps that provide effective support within the context of a child's problem-solving process, such benchmarks are still missing. Filling this gap is essential for enabling AI to deliver truly effective math tutoring.

We present MMTutorBench, the first multimodal benchmark for AI math tutoring. It evaluates MLLMs across diverse mathematical domains and education levels, comprising 770 problems centered on pedagogically significant key-steps where students often struggle. Each problem includes three tasks—Insight Discovery, Operation Formulation, and Operation Execution (Figure~\ref{fig:intro})—reflecting the stepwise nature of tutoring. Evaluation follows a rubric-guided framework scoring six fine-grained dimensions to ensure comprehensive assessment. We benchmark 12 leading MLLMs, revealing clear performance stratification between proprietary and open-source models and across tutoring-specific aspects. Beyond overall scores, we analyze input configurations: OCR-first pipelines degrade performance by losing spatial and diagrammatic cues, while few-shot prompting yields limited, model-dependent gains. Our rubric-based LLM-as-a-Judge also shows high inter-judge agreement, confirming evaluation reliability.
% ====

\vspace{-0.1cm}
\section{Related Work}
Our work is situated at the intersection of two active research areas: the application of Large Language Models (LLMs) in math tutoring and the development of multimodal mathematical reasoning capabilities. We review relevant literature in both domains to contextualize our contribution.

\subsection{LLMs in Math Tutoring}

Recent research has explored Large Language Models (LLMs) as scalable, personalized math tutors, primarily focusing on text-based dialogues. Studies have trained models to generate effective responses~\citep{Scarlatos_2025} and predict tutor strategies~\citep{ikram2025exploringllmspredictingtutor}. The focus has also extended to evaluation, with systems like MathTutorBench~\citep{Jakub2025MathTutor} using reward models to assess tutors on dimensions such as subject expertise and student understanding, while other work has used LLMs as evaluators for human tutors~\citep{thomas2025leveragingllmsassesstutor}. However, this text-centric paradigm overlooks critical real-world complexities. Existing multimodal systems have centered on affective dimensions, such as student emotion~\citep{kar2025mathbuddy}, rather than on the interpretation of visual mathematical content. Furthermore, studies confirm that an LLM's problem-solving proficiency does not equate to effective tutoring~\citep{gupta2025finalanswersevaluatinglarge}, and even state-of-the-art models remain prone to subtle reasoning errors~\citep{zhang2025mathematicalcomputationreasoningerrors}. To address the significant gap in visual-mathematical interpretation, MMTutorBench shifts the focus from purely textual dialogues to the multimodal task of interpreting a student's handwritten solution steps to provide effective feedback.
\vspace{-0.18cm}

\subsection{Multimodal Math Reasoning Benchmarks}

In parallel, the field of multimodal mathematical reasoning has advanced through key benchmarks designed for visually-presented problems. MathVista~\citep{mathvista} first establishes a foundational standard for comprehensive, reasoning-centric evaluation. MATH-Vision~\citep{mathvision} enhances problem difficulty and diversity by drawing from real math competitions, while MathVerse~\citep{mathverse} probes the depth of visual understanding by presenting problems in multiple variations. Shifting the focus from outcomes to the reasoning process itself, We-Math~\citep{wemath} pioneers fine-grained metrics for assessing principles like knowledge acquisition and generalization. However, these benchmarks are united by their singular focus on \textbf{problem-solving}. In contrast, MMTutorBench redefines the evaluation by assessing a model's ability to \textbf{act as a tutor}, a task centered on interpreting a student's handwritten intermediate steps to provide context-aware, scaffolded feedback.

\section{MMTutorBench}

\begin{figure*}[t]
    \centering
    \includegraphics[width=0.95\textwidth]{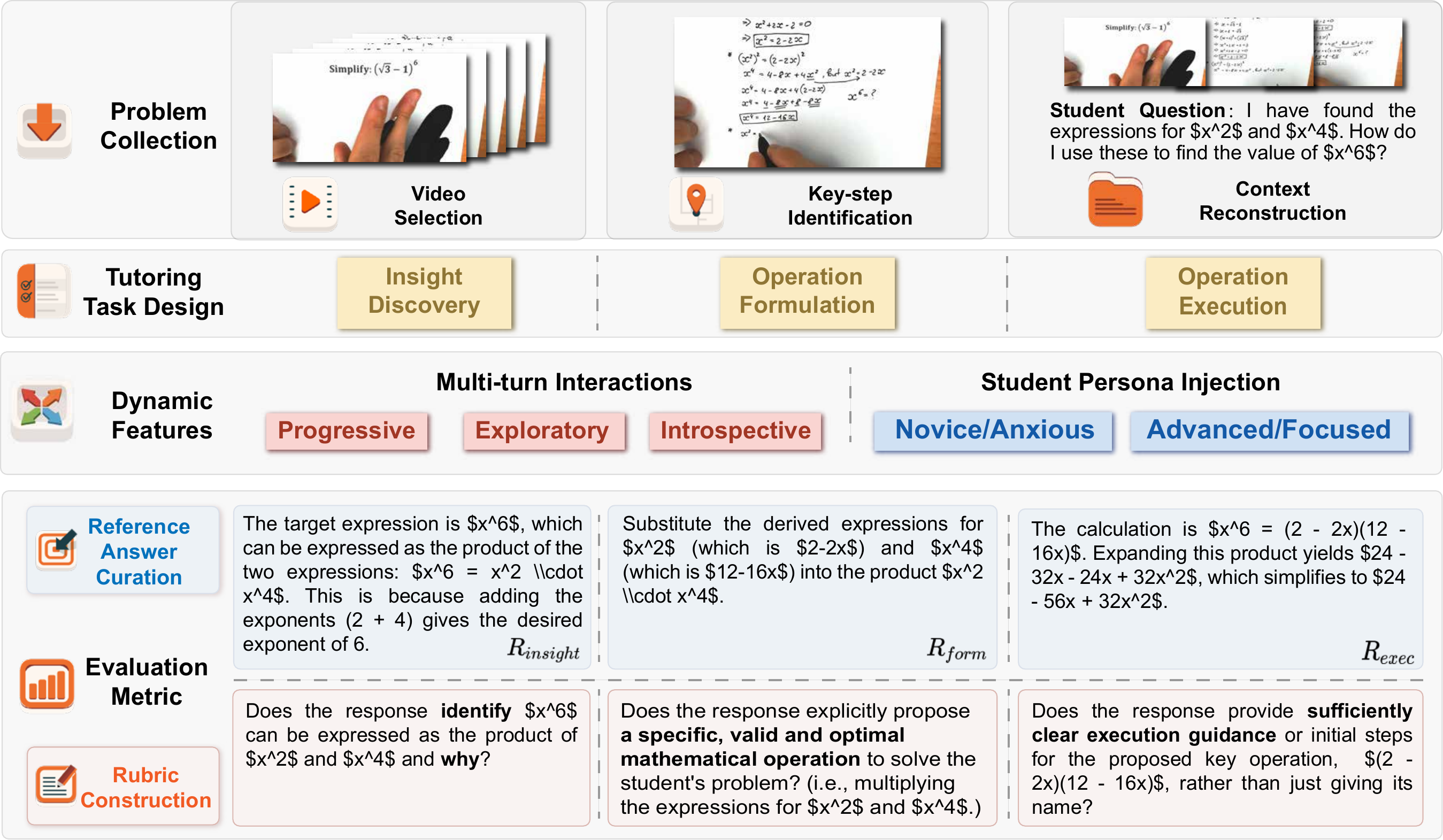}
    \vspace{-0.2cm}
    \caption{The data curation pipeline of MMTutorBench. We start by collecting problems including both images and questions. The model is instructed to fulfill 3 tutoring tasks for the input problem.}
    %The generated response are evaluated with carefully designed problem specific rubrics.}
    \label{fig:method} 
    \vspace{-0.5cm}
\end{figure*}

MMTutorBench is a comprehensive benchmark consisting of 770 carefully curated samples from real-world educational settings, designed to evaluate the tutoring capabilities of MLLMs.
In each sample, the model acts as a math tutor, interpreting students’ handwritten solutions and generating responses to guide them through challenging steps.

To construct these scenarios, we collect educational video frames and student-posed questions to simulate authentic handwritten problem-solving and help-seeking processes (\S~\ref{sec:input-prepare}).
We then decompose the tutoring objective into three task dimensions—Insight, Formulation, and Execution—following Pólya’s problem-solving principle~\cite{polya} (\S~\ref{sec:task-design}).
Finally, we design a rubric-guided evaluation framework that employs problem-specific rubrics for fine-grained, multi-perspective assessment of MLLM outputs (\S~\ref{sec:rubric-eval}).
Comprehensive statistics are provided in Table~\ref{tab:stats} and Appendix~\ref{appd:statistics}.

\begin{table}[h]
    \vspace{0.15cm}
    \centering
    \sisetup{
        group-separator={,}, % 使用逗号作为千位分隔符
        group-minimum-digits=4, % 超过4位才开始分组
        detect-weight, % 加粗命令也能作用于S列中的数字
    }
    \sisetup{group-separator={,}}
    \resizebox{\linewidth}{!}{
        \begin{tabular}{lr}
        \toprule
        \textbf{Statistic} & \textbf{Number} \\
        \midrule
        Total Problems & 770 \\
        Total Images & \num{1414} \\
        Images per Problem (1 / 2 / $\geq$3)  & 415  / 205 / 150  \\
        \midrule
        \multicolumn{2}{l}{\textbf{Question}} \\
        \quad Total Words & \num{233447} \\
        \quad Total /  Unique  Tokens & \num{330460} / \num{1342} \\
        \quad Avg. / Max. / Min. Tokens & 429.17 / 463 / 399 \\
        \midrule
        
        \multicolumn{2}{l}{\textbf{Reference Answer}} \\
        \multicolumn{2}{r}{\textit{Insight / OpForm. / OpExec. / Total}} \\
        \quad Total Words & \num{26794} / \num{15874} / \num{24375} / \num{67043} \\
        \quad Total Tokens & \num{40947} / \num{22344} / \num{47831} / \num{111122} \\
        \quad Unique Tokens & \num{1930} / \num{1433} / \num{128} / \num{3491} \\
        \quad Avg. Tokens & 53.2 / 29.0 / 62.1 / 144.3 \\
        \quad Max. Tokens & \num{123} / \num{89} / \num{192} / \num{404} \\
        \quad Min. Tokens & \num{21} / \num{7} / \num{13} / \num{41} \\
        \midrule
        
        \multicolumn{2}{l}{\textbf{Rubrics}} \\
        \quad Total Words & \num{262182} \\
        \quad Total /  Unique  Tokens & \num{363276} / \num{2547} \\
        \quad Avg. / Max. / Min. Tokens & 442.48 / 624 / 314 \\
        
        \bottomrule
        \end{tabular}
    }
    \vspace{-0.2cm}
    \caption{Statistics of MMTutorBench. The token number is counted by GPT-4o tokenizer. \texttt{Insight}, \texttt{OpForm.}, \texttt{OpExec.} are abbreviations for \texttt{Insight Discovery}, \texttt{Operation Formulation}, \texttt{Operation Execution}.}
    \vspace{-0.85cm}
    \label{tab:stats}
\end{table}

\subsection{Problem Collection}
\label{sec:input-prepare}
\paragraph{Video Selection.}
Mathematics educational videos that visually capture handwritten problem-solving processes with step-by-step explanations provide a natural foundation for constructing MMTutorBench. To this end, we curate a corpus of 292 high-quality instructional videos drawn from 14 mathematics-focused YouTube channels\footnote{\url{https://youtube.com}.}. The collection spans diverse mathematical domains (e.g., Algebra, Calculus) and educational stages ranging from junior and senior high school to university level.

\paragraph{Key-step Identification.}
Since mathematical problem solving involves multiple intermediate steps, we focus on the pedagogically critical ones—\textit{key-steps}, where learners often face confusion or require deeper reasoning. These typically involve applying a core theorem (e.g., Pythagorean theorem) or executing a pivotal algebraic operation (e.g., polynomial factoring). To identify them in tutoring videos, we use Gemini-2.5-Pro~\citep{gemini2.5report} to detect key-step timestamps, extract corresponding frames, and conduct manual quality checks. Further details appear in Appendix~\ref{sec:keyframe-extraction}.

\paragraph{Context Reconstruction.}
Educational videos are inherently dynamic (e.g., camera movement, page turning), which often causes crucial information—such as the problem statement or earlier steps—to move outside the frame. Thus, a single key-step frame may lack the broader context needed for comprehension. We first detect scene changes and extract representative frames. Human annotators then refine these frames by removing redundancy and filling gaps to ensure coherent context (details in Appendix~\ref{sec:context-reconstruction}).  Based on this contextualized representation, we then construct a tutoring question that simulates the inquiry a student would typically pose upon encountering difficulty at the key-step.

\subsection{Tutoring Task Design}
\label{sec:task-design}
With the specified visual and question inputs, we design tutoring-centered tasks that specify how models should respond. Rather than providing complete solutions, the tasks are structured to guide learners step by step, thereby cultivating transferable problem-solving skills. Inspired by Pólya’s problem-solving methodology~\citep{polya}, which frames reasoning as a staged process, including understanding the problem, devising a plan, and carrying out the plan, our benchmark operationalizes each stage at the level of a key-step through three tasks: 

\begin{itemize}[leftmargin=*, nosep, labelsep=0.5em]
    \item \textbf{[Insight Discovery]}
     demonstrates the ``\textit{why}'': the core principle or observation needed to make progress. It aims to help the student understand the underlying concept rather than only memorizing a procedure.
     
    \item \textbf{[Operation Formulation]}
     clarifies the ``\textit{what}'': the specific mathematical operation or concept that should be applied based on the key insight.
     
    \item \textbf{[Operation Execution]}
     explains the ``\textit{how}'': 
     the concrete execution of the prescribed operation, showing the immediate next step in the calculation without revealing the entire solution.
\end{itemize}

At inference, the tutor model receives the contextualized visuals, the task instructions, and optionally a student query, and completes the tasks sequentially (full prompt in Appendix~\ref{sec:prompts}).

\paragraph{Dynamic Interaction and Adaptability.} 
To simulate realistic educational dialogues beyond single-turn QA, we extend the task design into two advanced dimensions:
\begin{itemize}[leftmargin=*, nosep]
    \item \textbf{Multi-turn Interactions:} We strictly categorize student queries into three pedagogical levels—\textit{Progressive} (linear follow-up), \textit{Exploratory} (lateral clarification), and \textit{Introspective} (deep conceptual justification)—to test the model's ability to maintain scaffolding over time.
    \item \textbf{Student Persona Injection:} We introduce specific personas (e.g., \textit{Novice/Anxious} vs. \textit{Advanced/Focused}) via system prompts to evaluate whether models can dynamically adjust their tone and granularity.
\end{itemize}
% Detailed definitions and results analysis for these scenarios are provided in Appendix~\ref{sec:appendix_multiturn} and Appendix~\ref{sec:appendix_adaptivity}.

Detailed definitions and results analysis for these scenarios are provided in Appendix~\ref{sec:appendix_multiturn} and~\ref{sec:appendix_adaptivity}.

\subsection{Evaluation Metric}
\label{sec:rubric-eval}

\begin{table*}[t!]
    \centering
    \resizebox{0.9\textwidth}{!}{
     \begin{tabularx}{\textwidth}{@{} l l X @{}}
\toprule
\textbf{Category} & \textbf{Dimension} & \textbf{Evaluation Criteria (Condensed)} \\
\midrule
\multirow{2}{*}{\textbf{General}} 
 & \texttt{Brevity} 
 & Assesses whether the response is concise yet sufficient, avoiding redundancy while maintaining coverage comparable to the reference. \\
\cmidrule(l){2-3}
 & \texttt{Coherence} 
 & Assesses whether the response is logically consistent, factually accurate, and free of contradictions, relative to the reference. \\
\midrule
\multirow{4}{*}{\textbf{Specific}} 
 & \texttt{Insight Discovery} 
 & Examines whether the response identifies the key structure or observation required at this stage, consistent with \(R_{\text{insight}}\). \\
\cmidrule(l){2-3}
 & \texttt{Operation Formulation} 
 & Evaluates whether the response proposes the appropriate next conceptual operation, as indicated by \(R_{\text{form}}\). \\
\cmidrule(l){2-3}
 & \texttt{Operation Execution} 
 & Evaluates whether the response correctly and transparently performs the intended operation, as defined in \(R_{\text{exec}}\). \\
\cmidrule(l){2-3}
 & \texttt{Solution Scope Control} 
 & Checks whether the response remains focused on the current step, without advancing beyond \(R_{\text{insight}}, R_{\text{form}}, R_{\text{exec}}\). \\
\bottomrule
\end{tabularx}}
    \vspace{-0.2cm}
    \caption{Six-dimensional rubric for evaluating tutoring responses. Each dimension is operationalized relative to the step-specific reference answers \(R_{\text{insight}}, R_{\text{form}}, R_{\text{exec}}\).}
    \label{tab:rubric}
    \vspace{-0.5cm}
\end{table*}

Evaluating tutoring responses is challenging because the task is open-ended: there is no single “correct” answer that can be matched by accuracy or n-gram overlap. Traditional metrics such as BLEU~\citep{bleu} are thus inadequate. LLM-as-a-Judge methods offer a promising alternative, but naïvely applied they risk introducing bias and inconsistency~\citep{llmjudgebias}. To address this, we adopt a rubric-guided LLM-as-a-Judge framework, inspired by BiGGenBench~\citep{kim2025biggenbenchprincipledbenchmark}. The key idea is to anchor the evaluation of each sample to a problem-specific rubric, rather than relying on generic criteria. 

\paragraph{Reference Answer Curation.}
For each key-step sample we derive reference answers from the instructor’s explanation in the post-key-step content of the video. We deliberately extract only the content relevant to the immediate next step to ensure the evaluation focuses on the current tutoring step rather than the full solution. Based on this focused content, we construct three reference answers by human annotation and denote them as \(R_{\text{insight}}\) for Insight Discovery, \(R_{\text{form}}\) for Operation Formulation, and \(R_{\text{exec}}\) for Operation Execution.

\paragraph{Rubric Generation.}
With the reference answers, we define six task-specific rubric dimensions. The first two, \textit{Brevity} and \textit{Coherence}, capture general qualities of effective instructional text. The next three, \textit{Insight Discovery}, \textit{Operation Formulation}, and \textit{Operation Execution}, correspond directly to the structured tasks required by our benchmark. The final dimension, \textit{Solution Scope Control}, penalizes responses that provide the full solution instead of stepwise tutoring. A detailed explanation of each dimension and its scoring criteria is provided in Table~\ref{tab:rubric}.
The judge model evaluates candidate responses strictly against this rubric, rather than comparing them to references in a free-form way. This decomposition reduces the cognitive load on the judge model, improves consistency, and enables fine-grained assessment of both solution correctness and pedagogical effectiveness. The complete rubric is detailed in Table~\ref{tab:rubric}, with implementation details provided in Appendix~\ref{sec:rubric-generation}.

\section{Experiments}

We evaluate 12 leading MLLMs on our MMTutorBench to assess their capabilities 
in multimodal tutoring, investigate advanced tutoring scenarios spanning 
multi-turn interactions and student-level adaptability, study the impact of 
various input configurations through ablation, validate our LLM-as-a-Judge 
evaluation framework, and analyze the primary failure modes of the top-performing 
model.

\definecolor{ForestGreen}{RGB}{34,139,34}
\definecolor{IndianRed}{RGB}{205,92,92}

% up command (green)
\newcommand{\up}[1]{\textcolor{ForestGreen}{\tiny\textbf{$\uparrow$}#1}}
% down command (red)
\newcommand{\down}[1]{\textcolor{IndianRed}{\tiny\textbf{$\downarrow$}#1}}

\subsection{Experimental Setup}
To comprehensively evaluate our benchmark, we select 12 MLLMs, which span both proprietary and open-source models. 
Our evaluation suite includes 5 proprietary models: GPT-5~\citep{gpt5}, GPT-4o~\citep{openai2024gpt4o}, Gemini-2.5-Pro~\citep{gemini2.5report}, Gemini-2.0-Flash, and GPT-o3-2025-04-16. 
Additionally, we assess 7 leading open-source models: Qwen2.5-VL (7B-Instruct, 72B-Instruct)~\citep{Qwen2.5report}, InternVL3.5 (8B, 38B)~\citep{InternVL3.5report}, Gemma-3-27B-it~\citep{Gemma3report}, GLM-4.1V-9B-thinking~\citep{GLM4.1report}, and MiMo-VL-7B-RL~\citep{MiMoVLreport}. 
For all experiments, we employ a standardized prompt (detailed in Appendix~\ref{sec:prompts}) to ensure a fair comparison. Unless otherwise specified, our default experimental setting is zero-shot, where models are provided only with the task instruction and the relevant images, without any in-context examples or student query.
We assess responses using the rubric in Table~\ref{tab:rubric}, where each of the six criteria receives a binary score (0 or 1) for a maximum total score of 6.

\subsection{Main Results}

\begin{table*}[t!]
    \centering    
    \resizebox{0.83\linewidth}{!}{
        \begin{tabular}{l| c| c c c c c c}
            \toprule
            
            \textbf{Model} & \textbf{Tot.} & \textbf{Insight} & \textbf{OpForm.} & \textbf{OpExec.} & \textbf{Scope} & \textbf{Brevity} & \textbf{Coherence} \\
            \midrule
    
            \multicolumn{8}{c}{\textit{Proprietary Models}} \\
            \midrule
            
            Gemini-2.5-Pro & \textbf{4.69} & \textbf{0.79} & \textbf{0.73} & \textbf{0.73} & \textbf{0.69} & 0.78 & \textbf{0.97} \\
            GPT-5  & 4.32 & 0.76 & 0.70 & 0.70 & 0.36 & \textbf{0.83} & \textbf{0.97} \\
            GPT-o3 & 3.97 & 0.68 & 0.66 & 0.66 & 0.28 & 0.72 & \textbf{0.97} \\
            Gemini-2.0-Flash & 3.77 & 0.54 & 0.56 & 0.61 & 0.44 & 0.75 & 0.87 \\
            GPT-4o & 3.18 & 0.50 & 0.47 & 0.47 & 0.28 & 0.64 & 0.81 \\
            \midrule
    
            \multicolumn{8}{c}{\textit{Open-Source Models}} \\
            \midrule
    
            Qwen2.5-VL-72B-Instruct & \textbf{3.40} & \textbf{0.53} & \textbf{0.51} & \textbf{0.57} & 0.32 & 0.60 & 0.86 \\
            InternVL3.5-38B & 3.26 & \textbf{0.53} & 0.49 & 0.55 & 0.22 & 0.58 & \textbf{0.88} \\
            InternVL3.5-8B & 3.17 & 0.43 & 0.44 & 0.49 & 0.26 & \textbf{0.69} & 0.85 \\
            Gemma-3-27B & 2.87 & 0.38 & 0.37 & 0.39 & \textbf{0.35} & 0.66 & 0.72 \\
            MiMo-VL-7B-RL & 2.78 & 0.51 & 0.46 & 0.48 & 0.25 & 0.35 & 0.73 \\
            GLM-4.1V-9B & 2.55 & \textbf{0.53} & 0.50 & 0.54 & 0.12 & 0.11 & 0.75 \\
            Qwen2.5-VL-7B & 2.52 & 0.31 & 0.28 & 0.30 & \textbf{0.35} & 0.59 & 0.68 \\
                    
            \bottomrule
        \end{tabular}
    }
    \vspace{-0.2cm}
    \caption{Performance comparison of various models on our MMTutorBench. We report the total score (Tot.) and a detailed breakdown across six dimensions in our rubric. Column headers are abbreviations for: \texttt{Insight Discovery}, \texttt{Operation Formulation}, \texttt{Operation Execution}, \texttt{Solution Scope Control}, \texttt{Brevity}, and \texttt{Coherence}. }
    %For brevity, these dimension names will be abbreviated in subsequent tables and figures.
    \vspace{-0.4cm}
    \label{tab:main_results}
\end{table*}

Table~\ref{tab:main_results} summarizes the comprehensive evaluation results for all 12 models on MMTutorBench under the default setting. Our analysis of this data yields several key insights into the current capabilities and limitations of MLLMs on this challenging task.

\paragraph{A clear performance gap remains between proprietary and open-source models.}
As shown in Table~\ref{tab:main_results}, there is a clear distinction in performance between the two model categories. The leading proprietary model, Gemini-2.5-Pro, achieves a total score of 4.69. In contrast, the top-performing open-source model, Qwen2.5-VL-72B-Instruct, scored 3.40. This 1.29-point gap highlights that state-of-the-art proprietary systems still hold a considerable advantage in tackling the complex, multi-faceted reasoning required by our benchmark.

\paragraph{All models show a clear gap from human level.}
To establish an upper bound for performance, we evaluated human expert responses on a subset of the data (detailed in Table~\ref{tab:small-case}). The total human score reached 5.85, demonstrating a high standard of pedagogical quality. Even the most capable model in our evaluation, Gemini-2.5-Pro (4.69), remains more than a full point below this human baseline. This gap underscores the profound difficulty of the task and indicates that current MLLMs have not yet mastered the nuanced skills required for effective multimodal tutoring.
% This gap underscores the profound difficulty of the task and indicates that current MLLMs, despite their advancements, have not yet mastered the nuanced skills required for effective multimodal tutoring.

\paragraph{Tutoring Mode struggles with scope control.}
Intriguingly, models designed for specific educational  scenarios do not always excel on our benchmark. As shown in Table~\ref{tab:small-case}, the \textbf{GPT-4o Study Mode}~\cite{openai_study_mode}, tailored for learning applications, achieves a total score of 3.15, comparable to the standard GPT-4o’s 3.21. A closer look at the score breakdown shows a key trade-off: while the study mode may show competence in identifying insights (Insight: 0.62 vs. 0.53), it exhibits a severe deficiency in managing the answer's boundaries, scoring only 0.11 in \textbf{Solution Scope Control}. This failure to adhere to the problem's scope while attempting to be more explanatory demonstrates the benchmark's capacity to test not just correctness, but also crucial pedagogical skills like conciseness and focus. The poor performance of this specialized mode further validates the challenging and comprehensive nature of MMTutorBench.

\begin{table}[t!]
    \centering    
    \setlength{\tabcolsep}{3pt}
    \vspace{0.15cm}
    \resizebox{\linewidth}{!}{
    \begin{tabular}{l c c c c c c c}
        \toprule
        
        \textbf{Model} & \textbf{Tot.} & \textbf{Insight} & \textbf{OpForm.} & \textbf{OpExec.} & \textbf{Scope} & \textbf{Brevity} & \textbf{Coh.} \\
        \midrule
        
        Human & 5.85 & 0.97 & 0.97 & 0.97 & 0.97 & 0.98 & 0.98 \\
        \midrule

        GPT-4o & 3.21 & 0.53 & 0.45 & 0.51 & 0.29 & 0.62 & 0.80 \\
        \midrule
        
        GPT-Study & 3.15 & 0.62 & 0.47 & 0.53 & 0.11 & 0.51 & 0.91 \\     
        
        \bottomrule
    \end{tabular}
    }
    \vspace{-0.2cm}
    \caption{Performance comparison of human, GPT Study Mode, and GPT-4o on a 66-sample MMTutorBench subset.}
    \vspace{-0.35cm}
    \label{tab:small-case}
\end{table}

\subsection{Advanced Tutoring Capabilities}
\label{sec:exp_advanced}

Beyond single-step correctness, effective tutoring requires sustained scaffolding and pedagogical flexibility. We evaluate models on multi-turn consistency and persona adaptability using GPT-5 as a representative case study.

\paragraph{Multi-turn Scenarios.} 
The results reveal a significant inverse correlation between context length and scaffolding discipline. While GPT-5 maintains robust diagnostic accuracy across turns (\textit{Insight} score increases from 0.87 to 0.91), its ability to constrain the solution scope degrades sharply, with the \textit{Solution Scope Control} score dropping from 0.18 in Turn 2 to 0.09 in Turn 3. This deficiency is most pronounced in introspective queries requiring conceptual justification; in such cases, the scope control score collapses to 0.00, indicating that the model fails to withhold the final answer when pressed for deeper explanations. (See Appendix~\ref{sec:appendix_multiturn} for the complete performance dynamics across interaction types.)

\paragraph{Student-Level Adaptability.} 
The results highlight a substantial gap between problem-solving and tutoring capabilities: while GPT-5 achieves a high \textit{Insight} score of 0.72, its \textit{Adaptivity} score is disproportionately low at 0.30. This points to inherent behavioral rigidity, where models disregard prompted constraints on tone and granularity, reverting instead to their default, neutral training patterns regardless of the student's simulated needs. Full quantitative results and the rubric for adaptivity alignment are detailed in Appendix~\ref{sec:appendix_adaptivity}.

\subsection{Ablation Study on Input Variants}

\begin{figure}[t]
    \centering 
    \includegraphics[width=\linewidth]{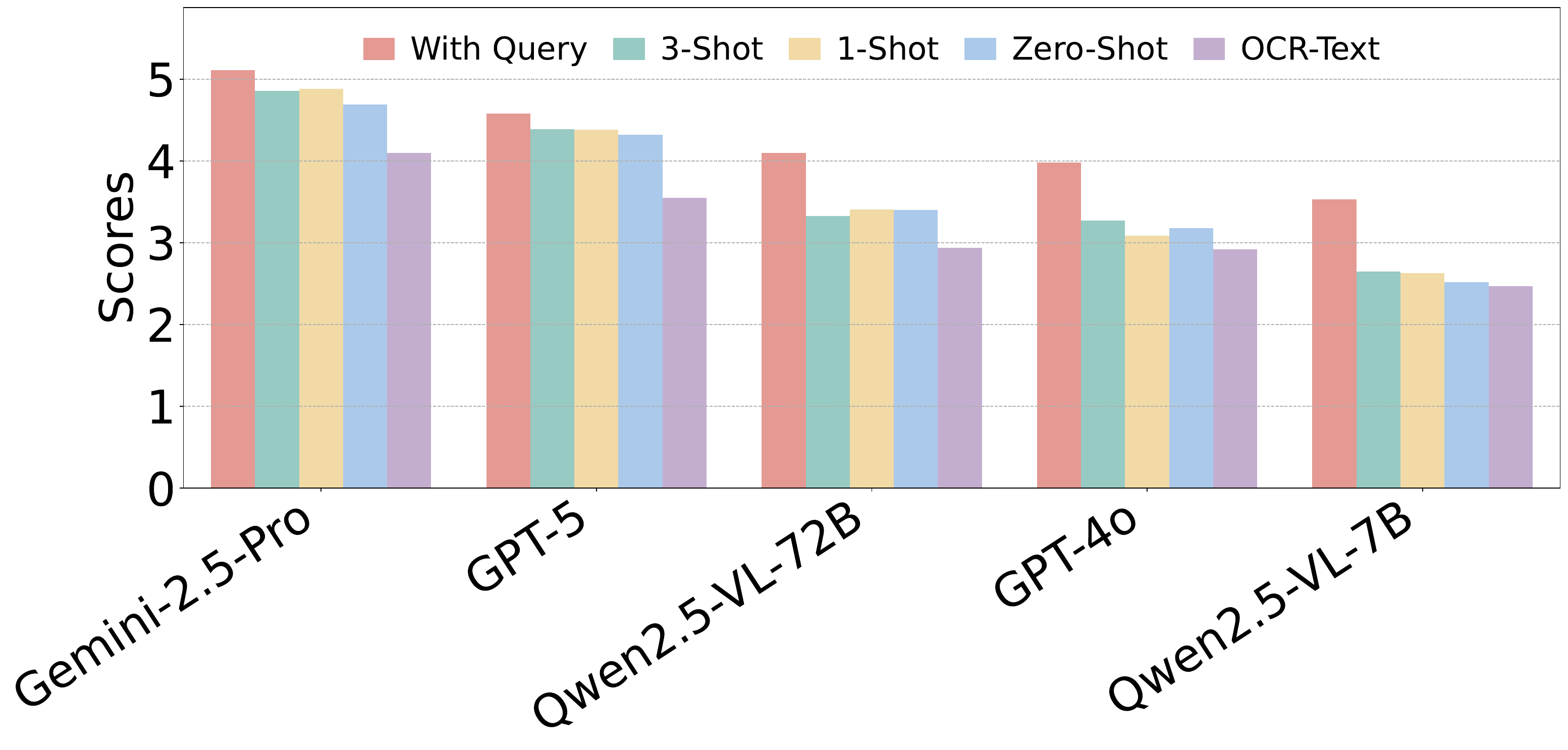}
    \vspace{-0.6cm}
    \caption{Performance comparison of various models across five distinct input variants. The variants include: (a) \textbf{Zero-Shot}, where only the images are provided; (b) \textbf{With Query}, which supplements the images with a corresponding textual student query; (c) \textbf{OCR-Text}, a pipeline approach where text is first extracted from the images via an OCR model and then fed to the language model; (d) \textbf{1-Shot} and (e) \textbf{3-Shot}, which provide one and three in-context examples, respectively. The results highlight the significant performance boost from including student queries and the critical limitations of the OCR-based pipeline approach.}
    \vspace{-0.4cm}
    \label{fig:input-variants}
\end{figure}

\paragraph{Impact of Few-Shot Prompts.}

% We first evaluate the models' in-context learning capabilities by comparing their performance in zero-shot, 1-shot, and 3-shot settings. As shown in Figure~\ref{fig:input-variants}, the inclusion of few-shot examples yields only marginal and model-dependent performance gains. For instance, the top-performing model, Gemini-2.5-Pro, demonstrates a modest improvement from 4.69 in the zero-shot setting to 4.86 in the 3-shot setting. This suggests that highly capable models can effectively leverage in-context examples to better align their responses with the desired pedagogical style. However, this trend is not universal. The performance of GPT-4o, for example, slightly decreases from 3.18 (zero-shot) to 3.09 (1-shot), indicating that for certain models, few-shot examples might inadvertently act as a constraint rather than a useful guide. This variability suggests that while in-context learning can offer a slight advantage for state-of-the-art models, the foundational reasoning capabilities of the model remain the dominant factor for success on our benchmark.

We evaluate in-context learning across zero-, 1-, and 3-shot settings. As shown in Figure~\ref{fig:input-variants}, few-shot prompting yields only marginal and model-dependent gains. While top-performing models, such as Gemini-2.5-Pro, improve slightly (4.69 to 4.86 in 3-shot), GPT-4o conversely exhibits performance degradation (3.18 to 3.09 in 1-shot). This variability suggests that while in-context learning can offer a slight advantage for state-of-the-art models, the foundational reasoning capabilities of the model remain the dominant factor for success on our benchmark.

\paragraph{Impact of Student Queries.}

% To assess how textual context influences multimodal reasoning, we compare model performance under two conditions: with only the images as input (Zero-Shot) versus supplementing the image with a textual student query. The results reveal a universal and substantial performance improvement across all evaluated models upon the inclusion of a student query. The gains are significant, with score increases ranging from 0.42 points for Gemini-2.5-Pro (4.69 to 5.11) to a remarkable 1.01 points for Qwen2.5-VL-7B (2.52 to 3.53). This strong, consistent trend suggests that the textual query serves as a powerful focusing mechanism. It helps models ground their visual analysis and precisely identify the student's point of confusion, thereby bypassing the more ambiguous and error-prone task of inferring the problem from visual context alone. This finding underscores the critical role of explicit, text-based cues in enabling effective multimodal tutoring.

To evaluate the impact of textual context, we compare performance in the image-only setting (Zero-Shot) versus the setting supplemented by textual student queries. The inclusion of student queries yields universal gains across all models, ranging from +0.42 for Gemini-2.5-Pro (4.69 to 5.11) to +1.01 for Qwen2.5-VL-7B (2.52 to 3.53). This trend suggests that textual queries serve as a powerful focusing mechanism to ground visual analysis, bypassing the more ambiguous and error-prone task of inferring students' confusion from visual context alone. This finding underscores the critical role of explicit, text-based cues in enabling effective multimodal tutoring.

\begin{table*}[t!]
    \centering
    \setlength{\tabcolsep}{3pt} 
     \resizebox{0.8\textwidth}{!}{
    \begin{tabular}{l ccccccc}
        \toprule
        \textbf{Model}  & \textbf{Tot.}  & \textbf{Insight} & \textbf{OpForm.} & \textbf{OpExec.} & \textbf{Scope} & \textbf{Brevity} & \textbf{Coherence}\\
        \midrule
        
        % 高分段模型: Gemini-2.5-Pro
        \textbf{Gemini-2.5-Pro} 
        & 4.77/4.87 & 0.80/0.80 & 0.74/0.74 & 0.74/0.74 & 0.71/0.71 & 0.80/0.93 & 0.98/0.94   \\

        % 高分段模型: GPT-5
        \textbf{GPT-5} 
        & 4.41/4.40 & 0.77/0.72 & 0.71/0.68 & 0.72/0.68 & 0.38/0.44 & 0.85/0.93 & 0.98/0.94   \\
        
        % 中分段模型: InternVL3.5-8B
        \textbf{InternVL3.5-8B} 
        & 3.19/3.10 & 0.43/0.43 & 0.45/0.40 & 0.50/0.48 & 0.27/0.29 & 0.68/0.80 & 0.86/0.74  \\

        % 低分段模型: MiMo-VL-7B-RL-2508
        \textbf{MiMo-VL-7B-RL} 
        & 2.75/2.78 & 0.51/0.46 & 0.46/0.44 & 0.48/0.46 & 0.26/0.26 & 0.34/0.59 & 0.72/0.58  \\

        \bottomrule
    \end{tabular}
    }
    \vspace{-0.2cm}
    \caption{
        Inter-judge reliability analysis on four representative models. 
        A random 90\% sample of the data is utilized for scoring comparison. Scores are presented in the format \textbf{GPT-o4-mini / Qwen3-30B-A3B-Instruct-2507}.}
        \vspace{-0.6cm}
    \label{tab:inter_judge_reliability}
\end{table*}

\paragraph{Impact of Modality: Image vs. OCR-Text.}

% Finally, to investigate the importance of direct visual processing, we compared the end-to-end multimodal approach with a pipeline method. In the latter, an OCR model, MiniCPM4.1-8B\citep{minicpm4}, first extracts text from the image, which is then processed by a large language model. This pipeline approach leads to a significant performance degradation across most models, confirming that direct visual analysis is indispensable. For example, the score of Gemini-2.5-Pro dropped from 4.69 (Zero-Shot) to 4.16 when relying solely on OCR-extracted text. This performance loss indicates that critical information is lost during the OCR process. Semantically crucial visual cues—such as the spatial layout of equations, diagrams, and non-textual symbols—are inadequately captured in a text-only representation. This finding validates that end-to-end visual understanding, rather than reasoning over extracted text, is essential for genuine multimodal comprehension in our benchmark.

To investigate the importance of direct visual processing, we compare the end-to-end multimodal approach with a pipeline method. In the latter, the powerful OCR model, MiniCPM4.1-8B\citep{minicpm4}, first extracts text from the image, which is then processed by the MLLM. This method causes significant performance degradation across most models—for instance, Gemini-2.5-Pro dropped from 4.69 to 4.16—confirming that direct visual analysis is indispensable. This indicates that semantically crucial visual cues—such as the spatial layout of equations, diagrams, and non-textual symbols—are inadequately captured in a text-only representation, validating that end-to-end visual understanding is essential for genuine multimodal comprehension in our benchmark.

\subsection{Rubrics Effectiveness}

% As established in our methodology (Section~\ref{sec:rubric-eval}), the complex nature of our benchmark necessitates an evaluation approach capable of assessing tutoring effectiveness and logical correctness. Consequently, we adopt a rubric-based methodology using an LLM-as-a-Judge. This section provides a rigorous validation of this chosen approach, focusing on two key aspects: its correlation with human judgment (validity) and its consistency across different evaluators (reliability).

To address the complex task of assessing tutoring effectiveness and correctness, we employ a rubric-based LLM-as-a-Judge. This section validates this methodology by examining two key aspects: its correlation with human judgment (validity) and consistency across evaluators (reliability).

% \begin{table}[t!]
%     \centering
%      \resizebox{0.85\linewidth}{!}{
%     \begin{tabular}{lcc}
%         \toprule
%         \textbf{Metric} & \textbf{Spearman ($\rho$)} & \textbf{Pearson ($r$)} \\
%         \midrule
%         BERTScore & 0.230 & 0.219 \\
%         BLEU & 0.233 & 0.267 \\
%         % ROUGE-1 & 0.380 & 0.493 \\
%         % ROUGE-2 & 0.375 & 0.391 \\
%         ROUGE-L & 0.341 & 0.386 \\
%         Standard Judge & 0.563 & 0.625 \\
%         \midrule
%         \textbf{Ours} & \textbf{0.652} & \textbf{0.725} \\
%         \bottomrule
%     \end{tabular}
%     }
%     \caption{Correlation coefficients of various metrics with human expert scores. The highest scores in each category are shown in bold.}
%     \vspace{-0.3cm}
%     \label{tab:correlation_table}
% \end{table}

\begin{table}[t!]
    \centering
    \vspace{0.2cm}
    \resizebox{\linewidth}{!}{
    \begin{tabular}{l l c c}
        \toprule
        \textbf{Category} & \textbf{Metric} & \textbf{Spearman ($\rho$)} & \textbf{Pearson ($r$)} \\
        \midrule
        
        \multirow{1}{*}{Embedding-based}
            & BERTScore & 0.230 & 0.219 \\
        \midrule
        
        \multirow{2}{*}{Rule-based}
            & BLEU & 0.233 & 0.267 \\
            & ROUGE-L & 0.341 & 0.386 \\
        \midrule
        
        \multirow{2}{*}{LLM-as-a-Judge}
            & Standard Judge & 0.563 & 0.625 \\
            & \textbf{Ours} & \textbf{0.652} & \textbf{0.725} \\
        
        \bottomrule
    \end{tabular}
    }
    \vspace{-0.2cm}
    \caption{Correlation with human expert judgments. Traditional rule-based and embedding-based metrics fail to capture pedagogical nuances, whereas our metric demonstrates superior alignment with expert scores.}
    \vspace{-0.6cm}
    \label{tab:correlation_table}
\end{table}

% To validate our methodology, we first benchmark it against human expert scores. As shown in Table~\ref{tab:correlation_table}, our rubric-based evaluation strongly correlates with human judgment (Pearson's $r=0.725$), significantly outperforming traditional NLP metrics and thus confirming its \textbf{validity}.

To validate our methodology, we first benchmark it against human expert scores. As shown in Table~\ref{tab:correlation_table}, our rubric-based evaluation strongly correlates with human judgment (Pearson's $r=0.725$), significantly outperforming all comparison baselines, thus confirming its validity.

We then establish the rubric's reliability through inter-judge analysis between \texttt{GPT-o4-mini} and \texttt{Qwen3-30B-A3B-Instruct-2507}~\citep{qwen3}. The scores demonstrated exceptionally high agreement, with a Pearson correlation exceeding 0.98 across the evaluated subset (visualized in Appendix~\ref{fig:scatter_plot}). This high degree of concordance, exemplified by nearly identical scores for models like GPT-5 (4.41 vs.\ 4.40), confirm that our evaluation is robust and minimizes judge-specific bias. For all main experiments, GPT-o4-mini is employed as the primary judge model.

\subsection{Error Analysis}

% [21.16883117 26.88311688 27.14285714 31.03896104 22.07792208  2.72727273]

\begin{figure}[h] 
    \centering 
    \includegraphics[width=0.45\textwidth]{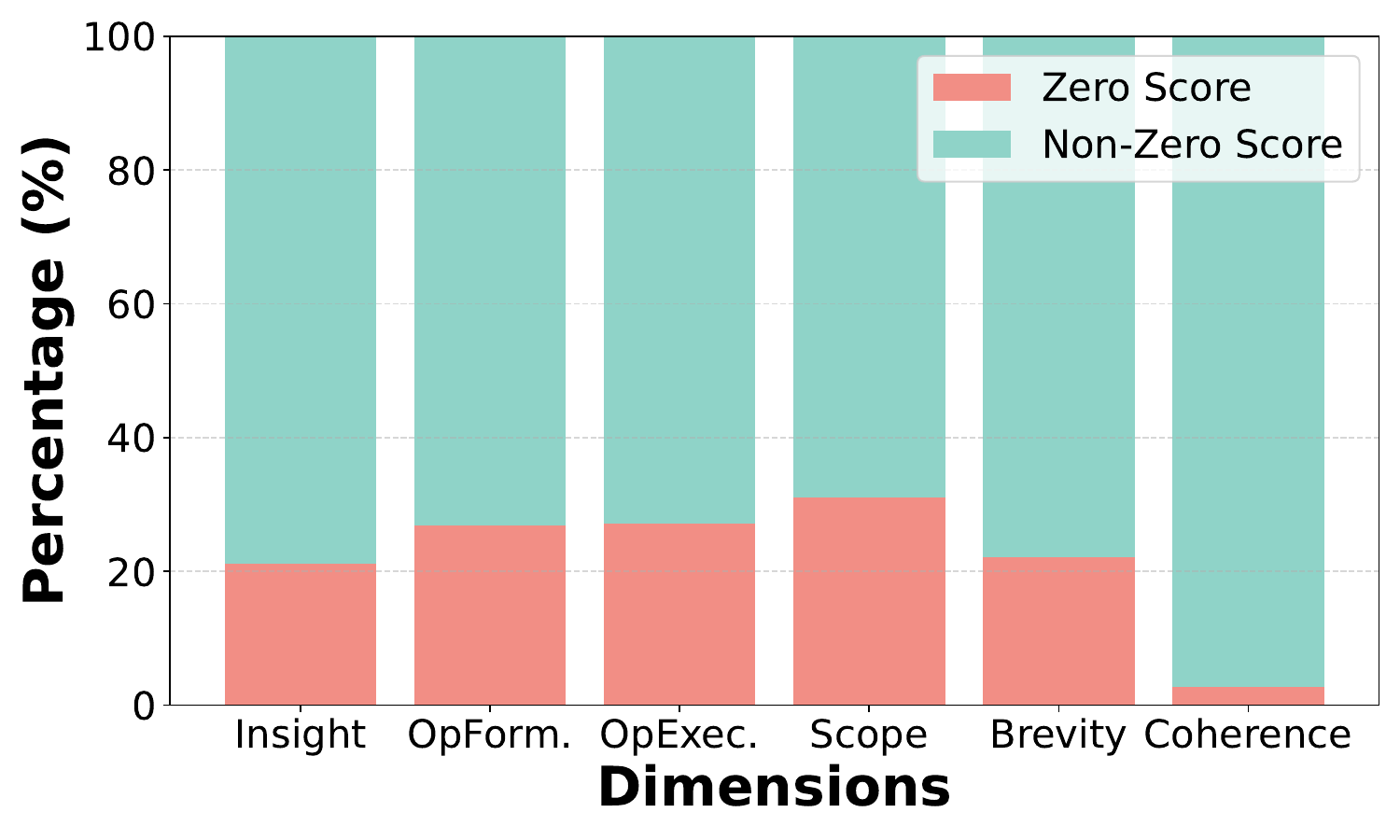}
    \vspace{-0.29cm}
    \caption{Error distribution for the top-performing model, Gemini-2.5-Pro. The chart displays the percentage of samples that received a score of zero in each of our six evaluation dimensions.} 
    \label{fig:error-analysis}
    \vspace{-0.554cm}
\end{figure}

To understand the primary failure modes, we conduct a detailed error analysis on the top-performing model, Gemini-2.5-Pro, by categorizing the instances where the model scored zero across our six dimensions. Figure~\ref{fig:error-analysis} presents the proportion of samples that failed in each dimension.

As illustrated, the most significant challenge for the model lies in Solution Scope Control, with nearly one-third (31.04\%) of its responses failing to adhere to the required scope of the solution. This is closely followed by failures in Operation Execution (27.14\%) and Operation Formulation (26.88\%). These three dimensions collectively indicate that while the model may identify a path forward, it struggles profoundly with correctly executing the necessary steps and constraining its output to the appropriate level of detail, often providing overly complex or irrelevant information.

% Furthermore, the model exhibits considerable weaknesses in Brevity (22.08\%) and Insight Discovery (21.17\%). This suggests that in about one-fifth of cases, the model either fails to produce a concise answer or misses the core insight required to solve the problem efficiently. These issues often compound the problems with operational control, leading to responses that are both incorrect and verbose.

Furthermore, the model exhibits considerable weaknesses in Brevity (22.08\%) and Insight Discovery (21.17\%), indicating that roughly one-fifth of responses lack conciseness or miss the core insight. These issues compound the operational failures above, yielding responses that are both incorrect and verbose.

A noteworthy finding, however, is the model's exceptional performance in Coherence. With a failure rate of only 2.73\%, the model's outputs are almost always logically structured, fluent, and easy to follow. This reveals a critical disparity: the model has mastered linguistic and structural coherence, but still lacks the deeper reasoning and self-control capabilities required for precise operational execution and scope management. The outputs are often well-formed but substantively flawed.

% To identify failure modes, we analyzed the zero-score distribution of Gemini-2.5-Pro (Figure~\ref{fig:error-analysis}). The primary bottleneck is \textit{Solution Scope Control} (31.0\% failure), followed closely by \textit{Operation Execution} (27.1\%) and \textit{Formulation} (26.9\%). These high failure rates indicate that while the model can often identify the path forward, it struggles profoundly with pedagogical constraints and precise calculation. Failures in \textit{Brevity} (22.1\%) and \textit{Insight} (21.2\%) further compound these issues. In contrast, the minimal failure rate in \textit{Coherence} (2.7\%) reveals a critical disparity: the model produces linguistically fluent but substantively flawed responses, lacking the necessary self-control for effective tutoring.
\section{Conclusion}
We introduce MMTutorBench, a comprehensive benchmark for evaluating Multimodal Large Language Models (MLLMs) on mathematical tutoring tasks. Our evaluation of 12 leading models reveals a significant performance gap between proprietary and open-source systems, with all models falling substantially short of human expert performance. We also show that direct visual grounding is indispensable, as text-only inputs are insufficient for effective tutoring. Furthermore, our findings indicate that while most models possess foundational visual understanding and problem-solving capabilities, they struggle to grasp the pedagogical concept of tutoring and often fail to appropriately control the scope of their guidance.

\section*{Limitations}
% We acknowledge two primary limitations. First, our benchmark evaluates single-turn, immediate feedback rather than the longitudinal effectiveness of a tutoring strategy over multiple sessions. Second, its scope is confined to English-language mathematics, which limits the generalizability of our findings across other subjects and languages. Future work could address these limitations by incorporating multi-step longitudinal scenarios and expanding the dataset to more subjects and languages.
We acknowledge two primary limitations. First, although our questions are pedagogically anchored in authentic video timestamps, they are simulated  and may not fully capture the linguistic ambiguity and spontaneity inherent in real-world learner interactions. Second, its scope is confined to English-language mathematics, which limits the generalizability of our findings across other subjects and languages. Future work could address these limitations by incorporating real-world classroom transcripts and expanding the dataset to more subjects and languages.

% \section{Ethical Considerations}
% \section*{Acknowledgments}

\newpage
\bibliography{custom}

\appendix

%%%%%%%%%%%%%%%%%%%%%%%%%%%5
\onecolumn
\newtcolorbox{promptbox}[2][]{
  enhanced,
  breakable,                
  colback=white,
  colframe=black,
  boxrule=1pt,
  arc=3mm,
  colbacktitle=black!70,
  coltitle=white,
  fonttitle=\sffamily\bfseries,
  fontupper=\small,
  title={#2},
  lower separated=false,
  segmentation style={draw=none},
  #1
}

\twocolumn
%%%%%%%%%%%%%%%%%%%%%%%%%%%

\section{Detailed Data Construction}
\subsection{Key-step Frame Extraction}
\label{sec:keyframe-extraction}

\begin{table}[t!]
    \centering
    \setlength{\tabcolsep}{3pt}
    \resizebox{\linewidth}{!}{
    \begin{tabular}{l c p{5.5cm}} 
        \toprule
        \textbf{Stage} & \textbf{Stats} & \textbf{Rejection Criteria} \\
        \midrule
        
        Video Sel. & $\sim 25.6\%$ & Videos are discarded due to poor handwriting legibility, low resolution, or lack of clear audio-visual alignment. \\
        \midrule 
        
        Key-Step Ext. & $\sim 63.4\%$ & Annotators filtered out LLM-suggested steps that are redundant, lack handwritten context, or involve pure calculation without pedagogical value. \\
        \midrule
        
        Rubric Ver. & \begin{tabular}{@{}c@{}} 100\% Ver. \\ 12.8\% Corr. \end{tabular} & Expert annotators manually verify all generated rubrics. Approximately 12.83\% require correction to remove factual errors to ensure strict alignment with the QA pairs. \\
        
        \bottomrule
    \end{tabular}
    }
    \caption{Statistics of the data construction pipeline. The rigorous filtering and verification process ensures high data quality. (Sel.: Selection, Ext.: Extraction, Ver.: Verification, Corr.: Corrected)}
    \label{tab:data-selection}
\end{table}

Our key-step frame extraction protocol is designed to systematically identify the most pedagogically valuable moments from each source video. The process is as follows:

\begin{enumerate}
    \item \textbf{Automated Candidate Generation:} We employ Gemini-2.5-Pro~\citep{gemini2.5report}, a powerful multimodal model, to analyze the full content of each video. Guided by a carefully crafted prompt, the model is instructed to identify pivotal steps in the problem-solving process.

    \item \textbf{Timestamp Pair Generation:} For each pivotal moment, the model outputs a \textbf{pair of precise timestamps} (in \texttt{HH:MM:SS} format) and a brief justification. This pair consists of the timestamp for the \textbf{critical step} itself and the timestamp for the \textbf{immediately preceding step}. To ensure our benchmark contains a diverse set of problems, we limit the extraction to a maximum of five such pairs per video.

    \item \textbf{Frame Pair Extraction:} The generated timestamp pairs are then used with the FFmpeg to extract the corresponding two static image frames for each identified moment.

    % \item \textbf{Manual Verification and Curation:} This is the most critical and labor-intensive phase of our data construction. Each extracted \textbf{pair of candidate frames} is subjected to a rigorous manual review by our annotators. This process involved assessing frames against strict criteria, including image clarity, pedagogical significance, and suitability for tutoring. Pairs that are blurry, redundant, or did not represent a truly pivotal moment are discarded. \textbf{Only the verified pairs are included in the final benchmark, ensuring the highest data quality.}

    \item \textbf{Human-in-the-Loop Verification and Curation:} This phase is the core of our quality control. As detailed in Table~\ref{tab:data-selection}, we applied strict rejection criteria at multiple stages:
    \begin{itemize}
        \item \textbf{Video Selection:} Prior to extraction, approximately \textbf{25.6\%} of raw videos are discarded due to poor legibility or audio-visual misalignment.
        \item \textbf{Step Filtration:} During the manual review of extracted frames, annotators filter out \textbf{63.4\%} of the LLM-suggested candidates. Steps are rejected if they are redundant, lack handwritten context, or involve pure calculation without pedagogical value.
        % \item \textbf{Rubric Correction:} Finally, expert annotators verified the generated rubrics. While 100\% are manually reviewed, \textbf{12.8\%} required correction to fix factual errors and ensure strict alignment with the QA pairs.
    \end{itemize}
    
    \item \textbf{Reliability Assessment:} To validate our annotation standards, we conduct a dual-blind annotation on 10\% of the data prior to the full-scale process. The experts achieve an \textbf{Inter-Annotator Agreement (IAA) of >90\%} on key-step identification, establishing a solid gold standard for the dataset creation.
\end{enumerate}

\begin{promptbox}{Prompt for Key-step Frame Extraction from Tutoring Videos}
    
    \textbf{Persona}
    \par
    You are a multimodal AI assistant analyzing a math tutoring video. Extract timestamps of key instructional frames that satisfy all of the following.
    
    \medskip\hrulefill\medskip
    
    \textbf{Task}
    \par
    Capture each \textbf{critical step} in the math tutoring video \textbf{and} the \textbf{immediately preceding step} before that critical step.
    \par
    \textbf{NOTE}
    \begin{enumerate}[leftmargin=*, topsep=2pt, itemsep=1pt]
        \item \textbf{Critical Step}: a pivotal stage in problem solving (e.g., equation transformation, formula introduction, concept explanation, etc.). 
        \item You must also capture the step just before the critical step, where the key equation is \textbf{about to appear but has not yet appeared}. The \textbf{previous step} and the \textbf{critical step} should be tightly connected; the only difference is whether the key equation is present. 
        \item Limit the total to \textbf{at most 5} pairs—choose the most representative moments. 
    \end{enumerate}
    
    \medskip\hrulefill\medskip

    \textbf{Requirement}
    \par
    \begin{enumerate}[leftmargin=*, topsep=2pt, itemsep=1pt]
        \item \textbf{Clear Handwriting}: The handwriting relevant to the step (equations, diagrams, etc.) is fully written and legible. 
        \item \textbf{Peak Clarity}: The handwriting is stable and complete—not mid-writing, blurry, or partially shown. 
        \item \textbf{Audio-Visual Match}: The narration clearly refers to the handwriting visible on screen. 
    \end{enumerate}

    \medskip\hrulefill\medskip

    \textbf{Exclude}
    \par
    \begin{itemize}[leftmargin=*, topsep=2pt, itemsep=1pt]
        \item Writing only the problem number/title. 
        \item Frames with blurry or incomplete handwriting. 
        \item Transition frames (e.g., board erasing). 
        \item Final boxed answers without explanation.
    \end{itemize}

    \medskip\hrulefill\medskip
    
    \textbf{Output Requirement}
    \par
    Return a \textbf{chronological list} of timestamp dictionaries with concise justifications.
    
    \medskip\hrulefill\medskip

    \textbf{Output Format}
    \begin{verbatim}
    {
      keyframe_timestamp: [MM:SS],
      prev_step_timestamp: [MM:SS],
      reason: 'xxx',
    }
    \end{verbatim}
    \emph{Note: These three fields denote the timestamp of the \textbf{critical step}, the timestamp of its \textbf{immediately preceding} step, and a brief justification for the key instructional moment, respectively.}
    
\end{promptbox}

\subsection{Context Reconstruction}
\label{sec:context-reconstruction}
Our context reconstruction protocol is designed to provide a comprehensive visual narrative leading up to each extracted key-step frame. Since a single frame often lacks the preceding information necessary for full comprehension (e.g., the original problem statement), this process segments the source video into a sequence of visually coherent images. The process is as follows:

\begin{enumerate}
    \item \textbf{Automated Scene Segmentation:} The process begins by programmatically identifying significant visual shifts in the video. We compute the \textbf{Structural Similarity Index (SSIM)}~\citep{ssim} score between every pair of consecutive frames. A potential scene boundary is flagged wherever this score drops below a threshold. To ensure that only meaningful transitions are retained and to filter out noise from transient motions (e.g., camera jitter), we apply a temporal filter that enforces a \textbf{minimum time interval} between consecutive boundaries.

    The formal algorithm is a two-step process. First, a set of candidate timestamps, $T_{\text{cand}}$, is identified:
    \begin{align*}
        T_{\text{cand}} &= \{t_n \mid \text{SSIM}(I_{n-1}, I_n) < \tau \}, \\
                        &\quad (\text{we use } \tau=0.8)
    \end{align*}
    Second, this candidate set is filtered iteratively to produce the final set of boundaries, $T_{\text{sb}}$, ensuring each boundary is separated by a minimum duration, $\Delta t_{\text{min}}$:
    \begin{align*}
      t_{s_1} &= t'_1 \\ 
              &\quad (\text{where } t'_1 \text{ is the earliest candidate}) \\[1ex] 
      t_{s_j} &= \min\{ t'_i \in T_{\text{cand}} \mid |t'_i - t_{s_{j-1}}| \ge \Delta t_{\text{min}} \}, \\ 
              &\quad \text{for } j > 1
    \end{align*}

    \item \textbf{Representative Frame Extraction:} Once the final set of scene boundaries is established, a single, clear \textbf{representative frame} is extracted from each resulting video segment. This transforms the video into an initial, condensed sequence of static images that summarizes the visual progression of the solution.

    \item \textbf{Manual Verification and Curation:} Similar to our key-step frame extraction, this phase is crucial for data quality. Our annotation team meticulously reviews the automatically generated sequence of representative frames. Their task is to refine this sequence by removing redundant images and supplementing any missing frames to repair logical or visual discontinuities. This meticulous curation ensures that the context provided for each key-step is a coherent and complete narrative.
\end{enumerate}

\subsection{Rubric Generation}
\label{sec:rubric-generation}

To enhance the accuracy and reliability of our evaluation, we developed problem-specific rubrics. The generation process for each sample's corresponding rubric is as follows:

\begin{enumerate}
    \item \textbf{Question-Answer Pair Extraction:} 
    Our process begins by analyzing the video transcripts. We first employ Gemini-2.0-Flash to process the subtitles and isolate the core mathematical problem-solving steps relevant to each key-step frame. Based on these extracted, concise solution steps, we then generate corresponding question-answer pairs. This output subsequently undergoes manual refinement, where annotators polish the questions to be clear and self-contained, and trim the answers to represent pedagogically meaningful steps.

    \item \textbf{Automated Rubric Generation:} 
    Based on each sample's question-answer pair and its full set of contextual images, we use Gemini-2.5-Pro with a structured prompt to generate scoring criteria for four specific dimensions: \textbf{Insight Discovery}, \textbf{Operation Formulation}, \textbf{Operation Execution}, and \textbf{Solution Scope Control}. These criteria are then combined with the requirements for two general dimensions (\textbf{Brevity} and \textbf{Coherence}) to create a preliminary six-dimensional rubric.
    
    \item \textbf{Manual Verification and Curation:} 
    The auto-generated rubrics undergo a rigorous manual verification process to ensure their precision and fairness. Our annotators meticulously review each scoring criterion, performing corrections, additions, or deletions as needed. The primary task is to ensure that the rubric's specific dimensions (\textbf{Insight Discovery}, \textbf{Operation Formulation}, \textbf{Operation Execution}) precisely and exclusively map to the corresponding components of the reference answer. This involves rephrasing ambiguous descriptions, clarifying conditions for earning points, and removing any criteria that do not directly pertain to the specific problem, thereby creating a highly reliable, problem-specific evaluation standard.
    While 100\% of the rubrics are manually reviewed, approximately 12.8\% required correction to fix factual errors and ensure strict alignment with the QA pairs.

\end{enumerate}

%%%%%%%%%%%%%%%%%%%%%%%%%%%%%%%%%%%%%%%%%
\begin{promptbox}{Prompt for Transcript Processing}
    
    You are helping clean up a noisy transcript from a math teaching video.
    Your task is to extract and return exactly \textbf{one} clear, concise sentence that conveys the core math explanation.
    
    \medskip
    \begin{itemize}[leftmargin=*, topsep=2pt, itemsep=1pt]
        \item Remove all filler words, background chatter, repetitions, or unrelated talk.
        \item Keep only mathematical explanation or reasoning.
        \item Do not include greetings, pauses, or commentary like 'so yeah', 'okay', 'um', etc.
        \item Make sure the sentence is standalone, complete, and easy to understand.
    \end{itemize}
    
    \medskip\hrulefill\medskip
    
    \textbf{[Noisy Transcript]}
    \par
    \texttt{\{text\}}
    
    \medskip
    
    \textbf{[Denoised Math Explanation (One Sentence)]}
    \par
    ...
    
    \end{promptbox}

\begin{promptbox}{Prompt for Generating Q\&A Pairs}
    
    \textbf{Role-Playing}
    \par
    You are a precise and logical tutor who guides students step by step through problem-solving. You do not need to solve the entire problem; you only need to instruct the student's question on the next key step and the reason for doing so. Your responses are purely \textbf{analytical and instructional}, with \textbf{no emotional tone} and \textbf{no conversational language}.
    
    \medskip
    \textbf{Task Description}
    \par
    Based on the learning context provided below, you need to generate two parts:
    \begin{enumerate}[leftmargin=*, topsep=2pt, itemsep=1pt, label=\arabic*.]
        \item \textbf{Student's Question:} Simulate a student who is trying to move forward in the problem but is confused about \textbf{what to do next}, based on the current step. The question should sound natural and authentic, using first-person language like "What should I do now?" or "How do I continue from here?" Notice: the question should NOT mention any specific mathematical operations or concepts when asking for the next step, but rather focus on the general direction or operation to take.
        \item \textbf{Tutor's Answer:} Provide a \textbf{precise and impersonal response} following the format:
        \begin{itemize}[leftmargin=*, topsep=2pt, itemsep=0pt]
            \item \textbf{[key detail]}: extract the key detail in the student's current state of the solution including the image and text, and explain the rationale for paying attention to this key detail.
            \item \textbf{[key operation]}: Based on the key detail, provide the very next critical operation the student should perform.
            \item \textbf{[next step]}: Perform the key operation in detail and determine the result.
        \end{itemize}
    \end{enumerate}
    
    \hrulefill
    
    \medskip
    \textbf{Key Example (Few-Shot)}
    
    \textbf{Image of current step: This is an example image of the current step of student's problem-solving process.}
    \texttt{'img:\{few\_shot\_img\_path\}'}
    
    \textbf{Context:}
    \begin{itemize}[leftmargin=*, topsep=2pt, itemsep=1pt]
        \item \textbf{Initial entire problem and student's current problem-solving stage:} "one plus a thousand to the power four plus a thousand and one to the power four divided by one plus a thousand squared plus a thousand and one squared. to simplify this fraction i'm going to let 1000 = x then this fraction becomes one plus x to power four and a thousand and one equals x plus one so it's x plus one all to the power four and this sum is divided by 1 plus x squared plus x plus 1 all squared. now i can simplify this algebraic expressionlet's do the easier one first which is 1 plus x squared plus x squared plus two x plus one. we can use the binomial theorem and we have x to the four and then it's 4 choose 1 x cubed 4 choose 2 x squared 4 choose 3 x and then it's just 1 at the end so i will not write 4 choose 4 but i'll just write 1 at the end. and we simplify we'll have 2 x to the power 4 plus 4 x cubed plus 6 x squared plus 4 x plus 2 at the top. we have 2 x squared plus 2x and then plus 2. and we realize that for the both the top and the bottom we have two so we'll cancel them out and we'll have x to the power four plus two x cubed plus three x squared plus two x plus one."
        \item \textbf{Student's current point of confusion:} "Three x squared plus two x plus one is the result."
        \item \textbf{Tutor's answer:} "The best solution is to factorize the numerator. If you cannot see how what do we get if we fracture this fraction if not then we can just use long division and you realize that with x squared on top. So it's 2x here not x and then we remove that we'll have x squared plus x plus one at the end so it turns out that the top is actually x squared plus this plus one all squared and then we can have a simplified expression and since x equals a thousand then the rest is very easy one million one thousand and one so this."
    \end{itemize}
    
    \textbf{Required Output:}
    \textbf{Student:} Okay, so I've simplified the top and bottom to get $x^4 + 2x^3 + 3x^2 + 2x + 1$ over $x^2 + x + 1$. I'm not really sure how to simplify this further. What should I do now?
    \par\medskip
    \textbf{Tutor:}
    \begin{itemize}[leftmargin=*, label={}, topsep=2pt, itemsep=1pt]
        \item \textbf{[key detail]}: We can find that the coefficients of the numerator polynomial $(x^4 + 2x^3 + 3x^2 + 2x + 1)$ are palindromic (or symmetric) and the numerator might be a multiple of the denominator, specifically its square.
        \item \textbf{[key operation]}: Thus we can consider factoring the numerator to attempt canceling the denominator and simplifying the fraction.
        \item \textbf{[next step]}: Factorize the numerator into $(x^2+x+1)(x^2+x+1)$. Then cancel the denominator $(x^2+x+1)$ from the numerator, resulting in a simplified expression of $(x^2+x+1)$.
    \end{itemize}

    \hrulefill

    \medskip
    \textbf{Your Task}
    
    \textbf{Image of current step:}
    \texttt{'img:\{img\_path\}'}
    
    \textbf{Context:}
    \begin{itemize}[leftmargin=*, topsep=2pt, itemsep=1pt]
        \item \textbf{Entire problem and student's current problem-solving stage:}
        \par
        \texttt{\{before\}}
        
        \item \textbf{Student's current point of confusion:}
        \par
        \texttt{\{sentence\}}
        
        \item \textbf{Tutor's answer:}
        \par
        \texttt{\{after\}}
    \end{itemize}
    
    \textbf{Required Output:}
    \textbf{Student:} ...
    \par
    \textbf{Tutor:} ...
    
\end{promptbox}

\begin{promptbox}{Prompt for Generating Rubrics}
    
    \noindent
    \textbf{Persona:} You are a meticulous AI assistant specializing in educational assessment. Your purpose is to craft precise, objective, and detailed evaluation rubrics. These rubrics are used to judge the quality of AI-generated hints for students solving math problems.
    
    \medskip
    
    \noindent
    \textbf{Goal} \\
    Your task is to generate a detailed, 4-point scoring rubric. This rubric will be used to evaluate a model’s response to a specific student’s question. You will create the rubric referenced on the provided question and its answer. Based on the output format example, you will fill in the “xxx” parts.
    
    \medskip
    
    \noindent
    \textbf{Note} \\
    When \texttt{condition\_for\_1} and \texttt{condition\_for\_0} parts of your rubric refer to specific information about the problem, it MUST be \textbf{decisive}; words like “such as” or “for example” that could mislead the scoring process are \emph{not} allowed.
    
    \medskip
    
    \noindent
    \textbf{Output format} \\
    Notice, the criteria and \texttt{id} in the \texttt{evaluation\_criteria} should be identical to the few-shot examples.  
    The rubric should be in JSON format, with the following structure:
    
    \begin{verbatim}
    {
        "task_description": "You are an AI 
        evaluator. Please assess an AI's 
        response to a student's math question 
        about xxx, based on the following
        `evaluation_criteria`. For each 
        criterion, assign a score of 0 or 1, 
        and summarize all scores in a single 
        JSON object.",
        "evaluation_criteria": [
            {
                "id": "insight_discovery",
                "criterion": "Does the response 
                identify and point out a key 
                structure or feature in the 
                expression that aids in solving 
                the problem?",
                "condition_for_1": "xxx",
                "condition_for_0": "xxx"
            },
            {
                "id": "operation_formulation",
                "criterion": "Does the response
                explicitly propose a specific, 
                valid and optimal mathematical 
                operation to solve the student's 
                problem?",
                "condition_for_1": "xxx",
                "condition_for_0": "xxx"
            },
            {
                "id": "operation_execution",
                "criterion": "Does the response 
                provide sufficiently clear 
                execution guidance or initial 
                steps for the proposed key 
                operation, rather than just 
                giving its name?",
                "condition_for_1": "xxx",
                "condition_for_0": "xxx"
            },
            {
                "id": "solution_scope_control",
                "criterion": "Is the response a 
                focused hint that only guides the 
                current step, rather than giving 
                a lengthy explanation or the full 
                answer? Note: If the response is 
                not a focused hint for the correct 
                step and is completely incorrect 
                or unhelpful, 
                the solution_scope_control 
                score should be 0, regardless of
     whether the condition_for_1 below is met. 
     If the response is a focused hint for the
     correct step, you still need to check if 
     it meets the condition_for_1 below.",
                "condition_for_1": "xxx",
                "condition_for_0": "xxx"
            }
        ],
        "output_format_instruction": "Please 
        strictly adhere to this JSON format for 
        the output:
     {\"insight_discovery\": <0|1>,
     \"operation_formulation\": <0|1>,
     \"operation_execution\":
     <0|1>, \"solution_scope_control\": <0|1>}"
    }
    \end{verbatim}

    \medskip
    
    \noindent
    \textbf{Few-shot examples:} \\
    \texttt{\{rubric\_few\_shot\}}
    
    \medskip
    
    \noindent
    Now, please generate a rubric for the following question and answer, using JSON format:
    
    \noindent
    \textbf{Question:} {question} \\
    \textbf{Reference Answer:} {answer} \\
    \textbf{Images:} `img:{img\_path}`, \textbf{Previous Images:} {', '.join(f"'img:{p}'" for p in prev\_img\_paths)}
    
    \medskip
    
    \noindent
    \texttt{Rubric:}

\end{promptbox}

\section{Prompts for Structured Output Generation}
\label{sec:prompts}

This section presents the exact prompts used to guide the model's generation process, ensuring full reproducibility of our experiments. We designed two prompt variants to handle distinct input conditions. The primary prompt, \textbf{Task Instruction without Student Query}, is used for tasks where the model must reason solely from the visual context of the provided images. The second, \textbf{Task Instruction with Student Query}, is an extension that incorporates an explicit student's question via the \texttt{\{question\}} placeholder.

Both prompts are structured to elicit a three-part response: `[Insight Discovery]`, `[Operation Formulation]`, and `[Operation Execution]`. They also include placeholders like \texttt{\{few\_shots\}} for injecting few-shot examples and \texttt{\{prev\_imgs\_str\}}/\texttt{\{kf\_img\_path\}} for the image inputs.

%%%%%%%%%%%%%%%%%%%%%%%%%%%%%%%%%%
\begin{promptbox}{Task Instruction w/o Student Query}
    % --- Role Definition ---
    You are a precise and logical tutor who guides students step by step through problem-solving.

    \medskip
    % --- Core Task ---
    \textbf{Task Description}
    \par
    A student is working on their math homework but got stuck after completing a few steps and does not know how to proceed. You will be given: 
    (1) a series of previous images in chronological order that show the original problem (the first image) and the student's step-by-step problem-solving process; 
    (2) a single current image that shows the student's current solution process; 
    based on the images, you need to identify the student's point of confusion and provide guidance on \textbf{the next key step} and \textbf{the detailed rationale for executing the key step}.
    
    \medskip
    \noindent
    \textbf{Note:} \textbf{the next key step} should be the single, most logical next step required to continue solving the problem. Your responses should be purely analytical and instructional, with no emotional tone and no conversational language. Your responses \textbf{MUST} be precise and concise. Do \textbf{NOT} include any unnecessary, overly long, or multi-step subsequent steps.
    
    \medskip
    \noindent
    Your response must include the following three parts:
    \begin{itemize}[leftmargin=*, topsep=2pt, itemsep=0pt]
    \item \textbf{Insight Discovery}: extract the key detail in the student's current state of the solution including the image and text, and explain the rationale for paying attention to this key detail.
    \item \textbf{Operation Formulation}: Based on the key detail, provide the very next critical operation the student should perform.
    \item \textbf{Operation Execution}: Perform the key operation in detail and determine the result.
    \end{itemize}
    
    \texttt{\{few\_shots\}}
    
    \medskip
    \textbf{Image}
    \begin{itemize}[leftmargin=*, topsep=2pt, itemsep=0pt]
    \item Previous images (chronological): \texttt{\{prev\_imgs\_str\}}
    \item Current image: \texttt{'img:\{kf\_img\_path\}'}
    \end{itemize}
    
    \medskip
    \textbf{Output}
    \par
    Use standard Latex notation for mathematical expressions. Your response \textbf{MUST} follow this format:
    \begin{itemize}[leftmargin=*, topsep=2pt, itemsep=0pt]
    \item \textbf{[Insight Discovery]}: xxx
    \item \textbf{[Operation Formulation]}: xxx
    \item \textbf{[Operation Execution]}: xxx
    \end{itemize}
    \end{promptbox}

%%%%%%%%%%%%%%%%%%%%%%%%%%%%%%%%%%%%%
\begin{promptbox}{Task Instruction w/ Student Query}
    \textbf{Role}
    \par
    You are a precise and logical tutor who guides students step by step through problem-solving.
    
    \medskip
    \textbf{Task Description}
    \par
    A student is working on their math homework but got stuck after completing a few steps and does not know how to proceed. You will be given: 
    (1) a series of previous images in chronological order that show the original problem (the first image) and the student's step-by-step problem-solving process; 
    (2) a single current image that shows the student's current solution process; 
    (3) a question that the student is asking about the next step;
    based on the images and question, you need to identify the student's point of confusion and provide guidance on \textbf{the next key step} and \textbf{the detailed rationale for executing the key step}.
    
    \medskip
    \noindent
    \textbf{Note:} \textbf{the next key step} should be the single, most logical next step required to continue solving the problem. Your responses should be purely analytical and instructional, with no emotional tone and no conversational language. Your responses \textbf{MUST} be precise and concise. Do \textbf{NOT} include any unnecessary, overly long, or multi-step subsequent steps.
    
    \medskip
    \noindent
    Your response must include the following three parts:
    \begin{itemize}[leftmargin=*, topsep=2pt, itemsep=0pt]
    \item \textbf{Insight Discovery}: extract the key detail in the student's current state of the solution including the image and text, and explain the rationale for paying attention to this key detail.
    \item \textbf{Operation Formulation}: Based on the key detail, provide the very next critical operation the student should perform.
    \item \textbf{Operation Execution}: Perform the key operation in detail and determine the result.
    \end{itemize}
    
    \texttt{\{few\_shots\}}
    
    \medskip
    \textbf{Image}
    \begin{itemize}[leftmargin=*, topsep=2pt, itemsep=0pt]
    \item Previous images (chronological): \texttt{\{prev\_imgs\_str\}}
    \item Current image: \texttt{'img:\{kf\_img\_path\}'}
    \end{itemize}
    
    \medskip
    \textbf{Student's question}
    \par
    \texttt{\{question\}}
    
    \medskip
    \textbf{Output}
    \par
    Use standard Latex notation for mathematical expressions. Your response \textbf{MUST} follow this format:
    \begin{itemize}[leftmargin=*, topsep=2pt, itemsep=0pt]
    \item \textbf{[Insight Discovery]}: xxx
    \item \textbf{[Operation Formulation]}: xxx
    \item \textbf{[Operation Execution]}: xxx
    \end{itemize}
    \end{promptbox}

%%%%%%%%%%%%%%%%%%%%%%%%%%%%%%%%%%%%%%%%%%%

\section{Evaluation Validation}
\label{sec:evaluation_validation} 

\subsection{Inter-Judge Reliability}

% 再画个散点图说明在全部模型范围的相关性系数很高，挑选了三个分数段：高、中、低表格对比
% 1. 说明 rubric 设计很好
% 2. 说明rubric设计能够减少对 judge 模型参数量、推理能力的依赖
\begin{figure}[h!]
    \centering
    \includegraphics[width=0.45\textwidth]{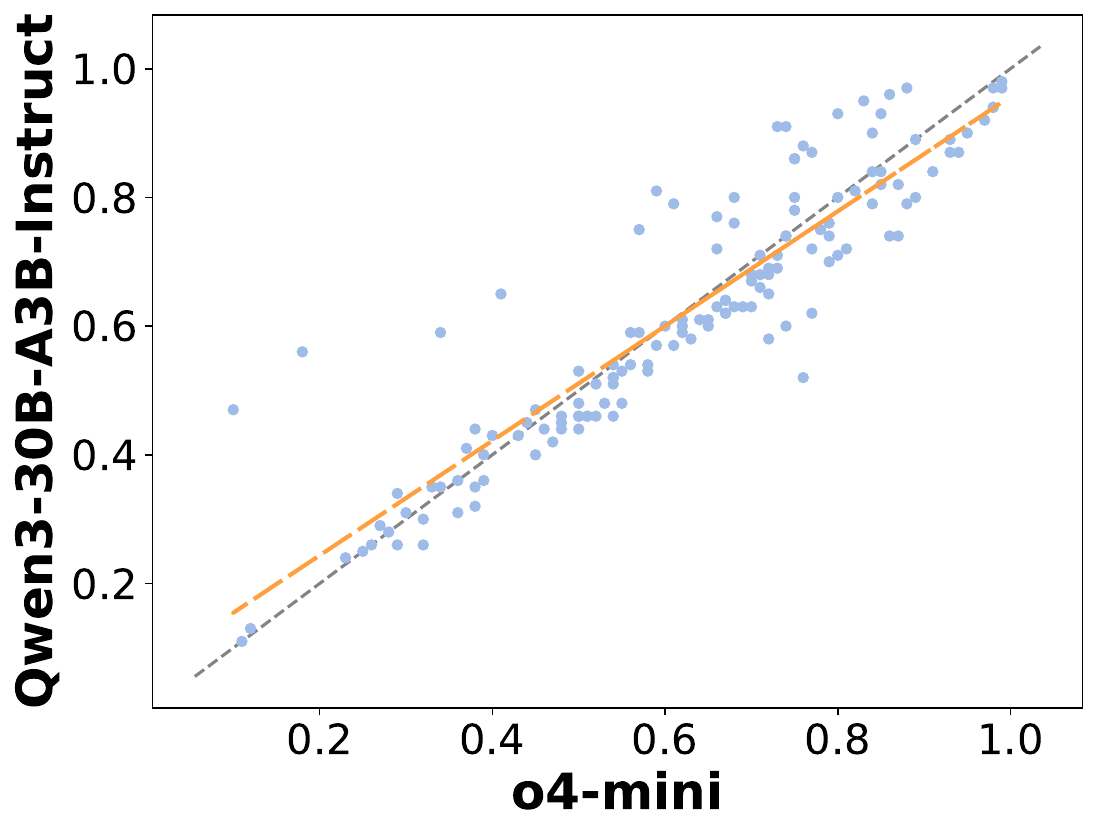}
    \caption{Inter-judge reliability of our evaluation rubric. The plot shows the correlation between average scores assigned to all 12 MLLMs by two independent judge models: GPT-o4-mini and Qwen3-30B-A3B-Instruct-2507.}
    \label{fig:scatter_plot}
\end{figure}

To supplement our inter-judge reliability analysis in the main text, Figure~\ref{fig:scatter_plot} provides a scatter plot of the average scores assigned to all 12 MLLMs by the two independent judge models, demonstrating a high alignment between evaluation scores from smaller, open-source reasoning models and their more powerful, proprietary counterparts. 

Furthermore, Table~\ref{tab:inter_judge_reliability} reveals that our meticulously crafted, problem-specific criteria result in higher scoring consistency for the four specific dimensions across models of varying capabilities. In contrast, the two general dimensions (Brevity and Coherence), which use uniform criteria for all samples, show a greater but still acceptable variance in scores. This highlights the effectiveness of our problem-specific rubric design and the robustness of our overall evaluation framework.

\subsection{Extended Human Correlation Analysis}
\label{subsec:human_correlation}

While expert human annotation for multimodal mathematical tutoring is highly labor-intensive, validating our automated metric across diverse domains and difficulty levels is crucial to ensure its statistical robustness. To achieve this, we expanded our human evaluation set by sampling 100 new instances across four distinct domains: Algebra, Analysis \& Calculus, Geometry, and Statistics \& Probability Theory. Note that these four domains correspond to a finer subdivision of the benchmark's three meta-categories (defined in Appendix~\ref{appd:statistics}), where \textit{Advanced} is here split into \textit{Analysis \& Calculus} and \textit{Statistics \& Probability Theory} for more granular validation. To rigorously test metric reliability under challenging conditions, we intentionally over-sampled hard problems (difficulty scores 4--5), which are under-represented in the full benchmark (5.6\%) but constitute 66\% of this evaluation subset. This design ensures the human correlation analysis is not driven solely by easier problems for which both automatic metrics and humans tend 
to agree trivially. We then computed the Pearson correlation ($r$) between human expert ratings and two evaluation methods: our rubric-based LLM-as-a-judge and a standard (rubric-free) LLM-as-a-judge baseline.

As illustrated in Table~\ref{tab:human_corr_domain}, our automated metric maintains a high and consistent correlation with human judgment across all mathematical sub-domains. It achieves an overall correlation of 0.8926, significantly outperforming the standard judge's 0.4487.

Crucially, given the intentional over-sampling of hard problems in this subset, we further stratified the correlation analysis by difficulty level (Table~\ref{tab:human_corr_difficulty}) 
to verify that the metric performs reliably across the full difficulty spectrum, not just on the dominant hard cases. The results demonstrate that our rubric-based method maintains robust alignment even on the most challenging problems (Pearson $r > 0.80$ for difficulty levels 4 and 5). In contrast, the standard judge completely fails on complex tasks, with its correlation dropping to negative or near-zero values. These findings conclusively prove that our problem-specific rubric design is essential for effectively aligning LLM evaluations with the nuanced pedagogical judgments of human experts, regardless of the mathematical domain or problem difficulty.

\begin{table}[h!]
\centering
\small
\resizebox{\columnwidth}{!}{
\begin{tabular}{lccc}
\toprule
\textbf{Domain} & \textbf{Count} & \textbf{Rubric-based (Ours)} & \textbf{Standard Judge} \\
\midrule
Algebra & 34 & 0.8419 & -0.0759 \\
Analysis \& Calculus & 33 & 0.7925 & 0.2763 \\
Geometry & 16 & 0.9492 & 0.4846 \\
Stat. \& Prob. Theory & 17 & 0.9722 & 0.4211 \\
\midrule
\textbf{Total} & \textbf{100} & \textbf{0.8926} & \textbf{0.4487} \\
\bottomrule
\end{tabular}
}

\caption{Pearson Correlation across Mathematical Domains. The four domains here correspond to a finer split of the benchmark's \textit{Advanced} meta-category (Analysis \& Calculus and Stat. \& Prob. Theory) and Algebra and Geometry. Our rubric-based metric maintains consistently high correlation with human expert ratings across all sub-domains.}
\label{tab:human_corr_domain}
\end{table}

\begin{table}[h!]
\centering
\small
\resizebox{\columnwidth}{!}{
\begin{tabular}{lccc}
\toprule
\textbf{Difficulty Score} & \textbf{Count} & \textbf{Rubric-based (Ours)} & \textbf{Standard Judge} \\
\midrule
$\le 2$ & 20 & 0.9612 & 0.5914 \\
$3$ & 14 & 0.9756 & 0.4715 \\
$4$ & 17 & 0.9188 & -0.1667 \\
$5$ & 49 & 0.8013 & 0.1564 \\
\midrule
\textbf{Total} & \textbf{100} & \textbf{0.8926} & \textbf{0.4487} \\
\bottomrule
\end{tabular}
}
\caption{Pearson Correlation across Problem Difficulty. Difficulty scores follow a 1--5 scale, where $\le$2 = Easy, 3 = Medium, and 4--5 = Hard (as defined in Appendix~\ref{appd:statistics}). Our method maintains robust alignment with human judgments even on the most challenging problems (levels 4 and 5), where the standard judge 
fails.}
\label{tab:human_corr_difficulty}
\end{table}

\section{Evaluation of Student-Level Adaptability}
\label{sec:appendix_adaptivity}

To further assess the pedagogical capabilities of MLLMs beyond mere problem-solving, we conduct an additional experiment focusing on \textbf{Student Adaptivity}. This experiment evaluates whether models can dynamically adjust their explanatory tone and granularity based on specific student personas defined in the system prompt.

\subsection{Experimental Setup}
We introduce a \textbf{Persona-Based Injection} protocol. For each problem in the benchmark, we condition the model with one of two distinct student metadata profiles via the system prompt. The detailed instructions and constraints for each persona are contrasted in Table~\ref{tab:persona_prompts}.

\begin{table*}[t!]
    \centering
    \small
    \renewcommand{\arraystretch}{1.3}
    \resizebox{\textwidth}{!}{
    \begin{tabular}{p{0.15\textwidth} p{0.4\textwidth} p{0.4\textwidth}}
        \toprule
        \textbf{Dimension} & \textbf{Persona A: Novice/Anxious} & \textbf{Persona B: Advanced/Focused} \\
        \midrule
        
        \textbf{Student Metadata} & 
        \begin{itemize}[leftmargin=*, nosep]
            \item \textbf{Proficiency:} Novice
            \item \textbf{State:} Anxious/Frustrated
            \item \textbf{Goal:} Confidence building
        \end{itemize} & 
        \begin{itemize}[leftmargin=*, nosep]
            \item \textbf{Proficiency:} Advanced
            \item \textbf{State:} Neutral/Hurried
            \item \textbf{Goal:} Efficiency \& Key Tricks
        \end{itemize} \\
        \midrule
        
        \textbf{Tone Instruction} & 
        Be \textbf{warm, encouraging, and supportive}. Use phrases like "You're doing great" or "Don't worry" to lower anxiety. & 
        Be \textbf{direct, professional, and concise}. Avoid filler words or emotional support; treat the student as a peer. \\
        \midrule
        
        \textbf{Granularity} & 
        Break down the execution into \textbf{very simple, explicit micro-steps}. Do \textbf{NOT} skip any intermediate calculation. & 
        \textbf{Skip trivial arithmetic} or algebraic manipulations. Focus only on non-trivial transformations. \\
        \midrule
        
        \textbf{Strategy} & 
        Explain the basic concept behind the insight \textbf{patiently}, assuming no prior intuition. & 
        Highlight the core insight or \textbf{mathematical trick immediately}. Provide a quick hint rather than a lecture. \\
        
        \bottomrule
    \end{tabular}
    }
    \caption{System Prompt Instructions for Student Personas. We inject these specific constraints into the system prompt to evaluate the model's ability to adapt its pedagogical style.}
    \label{tab:persona_prompts}
\end{table*}

\subsection{Evaluation Metric}
We employ an LLM-as-a-Judge approach using \texttt{o4-mini} to quantify adaptability. Unlike standard correctness metrics, this metric focuses purely on pedagogical fit. We also introduce a binary rubric dimension, \textbf{Adaptivity Alignment}, as defined in Table~\ref{tab:adaptivity_rubric}.

\begin{table}[h]
    \centering
    \small
    \renewcommand{\arraystretch}{1.2}
    \resizebox{\linewidth}{!}{
    \begin{tabular}{p{0.95\linewidth}}
        \toprule
        \textbf{Metric: Adaptivity Alignment (0 or 1)} \\
        \midrule
        \textbf{Criterion:} Does the tutor's response explicitly adapt its tone, granularity, and strategy to align with the specific \textsc{Student Persona} constraints provided in the instructions? \\
        \midrule
        \textbf{Score 1 (Strict Alignment):} \\
        The response successfully embodies the required persona: \\
        $\bullet$ \textit{For Novice:} The tone is encouraging AND the explanation is detailed without skipping steps. \\
        $\bullet$ \textit{For Advanced:} The tone is concise/direct AND trivial steps are omitted to focus on the core insight. \\
        The response follows the specific formatting or stylistic constraints requested. \\
        \midrule
        \textbf{Score 0 (Generic/Mismatched):} \\
        The response fails to adapt or contradicts the persona: \\
        $\bullet$ Uses a generic, robotic tone regardless of the student's state. \\
        $\bullet$ Is mismatched (e.g., overly verbose/pedantic for an Advanced student, or too abstract for an Anxious Novice). \\
        $\bullet$ Ignores specific instructions regarding granularity (e.g., skipping steps when asked not to). \\
        \bottomrule
    \end{tabular}
    }
    \caption{Rubric for Adaptivity Alignment. This binary metric evaluates style and pedagogical fit, independent of mathematical correctness.}
    \label{tab:adaptivity_rubric}
\end{table}

We compare this adaptability score against the standard \textbf{Insight Score}, which measures the mathematical correctness and visual understanding of the solution.

\subsection{Results and Analysis}
We evaluate two representative models, GPT-5 and Qwen2.5-VL-72B-Instruct. The results are summarized in Table~\ref{tab:student-level-adaptive}.

\begin{table}[h]
    \centering
    \setlength{\tabcolsep}{3pt}
    \resizebox{0.85\linewidth}{!}{
    \begin{tabular}{l c c} 
        \toprule
        \textbf{Model} & 
        \begin{tabular}{@{}c@{}}\textbf{Insight Score} \\ \small{(Math Comp.)}\end{tabular} & 
        \begin{tabular}{@{}c@{}}\textbf{Adaptivity Score} \\ \small{(Pedagogical Fit)}\end{tabular} \\
        \midrule
        
        GPT-5 & 0.72 & 0.30 \\
        Qwen2.5-VL-72B-Instruct & 0.51 & 0.27 \\
        
        \bottomrule
    \end{tabular}
    }
    \caption{Evaluation of student-level adaptivity. The significant gap between Insight and Adaptivity scores reveals that strong mathematical competence does not inherently translate into effective pedagogical flexibility.}
    \label{tab:student-level-adaptive}
\end{table}

\paragraph{Solving $\neq$ Tutoring.} 
A critical finding from this experiment is the divergence between problem-solving capability and tutoring adaptability. As shown in Table~\ref{tab:student-level-adaptive}, while \textbf{GPT-5} demonstrates superior mathematical competence (Insight of 0.72), it performs poorly in Adaptivity (0.30). 

Qualitative analysis reveals that despite explicit system instructions, stronger models often suffer from \textit{behavioral rigidity}, prioritizing their default training preference for ``standard solutions'' over the specific pedagogical needs of the user. This result underscores the unique value of our benchmark: it highlights that a strong ``Math Solver'' is not necessarily a capable ``Adaptive Tutor'', pointing to a crucial direction for future alignment research in educational AI.

\section{Weighted Evaluation Framework}
\label{sec:weighted_eval}

In our main evaluation, we report unweighted averages across all rubric dimensions. However, not all dimensions are equally critical for effective tutoring. For instance, correctly diagnosing a student's misconception (Insight) is arguably more fundamental than the conciseness of the response (Brevity).

To validate the robustness of our benchmark, we introduce a \textbf{Weighted Evaluation Framework}. As detailed in Table~\ref{tab:weight_rationale}, we assign distinct weights to each dimension to reflect their pedagogical priority.

\subsection{Rationale for Weight Assignment}
We categorize the dimensions into four priority levels based on educational taxonomy:
\begin{enumerate}
    \item \textbf{Critical (25\%):} \textit{Insight Discovery}. This is the "soul" of tutoring. Without accurate diagnosis of the mathematical structure, tutoring is impossible.
    \item \textbf{High (20\%):} \textit{Solution Scope Control}. This distinguishes a "tutor" from a "solver." It ensures the model scaffolds the learning rather than revealing the final answer.
    \item \textbf{Standard (15\% each):} \textit{Coherence, Operation Formulation, Operation Execution}. These represent the baseline correctness and methodological soundness.
    \item \textbf{Secondary (10\%):} \textit{Brevity}. While concise language reduces cognitive load, it is secondary to factual accuracy and pedagogical strategy.
\end{enumerate}

% --- 权重定义的详细表格 ---
\begin{table}[h]
    \centering
    \setlength{\tabcolsep}{4pt}
    \resizebox{\linewidth}{!}{
    \begin{tabular}{l c p{8cm}} % 使用 p{8cm} 自动换行
        \toprule
        \textbf{Dimension} & \textbf{Weight} & \textbf{Pedagogical Rationale} \\
        \midrule
        
        \textbf{Insight Discovery} & \textbf{0.25} & \textit{Diagnosis Capability.} It serves as the foundation of scaffolding, requiring the model to identify the deep mathematical structure rather than just calculating numbers. \\
        \midrule
        
        \textbf{Solution Scope Control} & \textbf{0.20} & \textit{Pedagogical Pacing.} Critical for preventing "spoilers." It forces the model to guide the student step-by-step rather than outputting the final result immediately. \\
        \midrule
        
        \textbf{Coherence} & 0.15 & \textit{Reliability Baseline.} In math tutoring, tolerance for hallucinations or contradictions is near zero. Factual errors negate all educational value. \\
        \addlinespace[3pt]
        
        \textbf{Op. Formulation} & 0.15 & \textit{Methodology.} Bridging "Insight" and "Execution" by explicitly stating the correct strategic path (e.g., "Use factorization"). \\
        \addlinespace[3pt]
        
        \textbf{Op. Execution} & 0.15 & \textit{Demonstration.} While important, "pointing the way" (Formulation) is often more pedagogically valuable than "doing the math" (Execution) for the student. \\
        \midrule
        
        \textbf{Brevity} & 0.10 & \textit{User Experience.} Concise responses lower cognitive load, but this is a "nice-to-have" quality compared to correctness and pedagogical validity. \\
        
        \bottomrule
    \end{tabular}
    }
    \caption{\textbf{Pedagogical weight assignment.} Weights are distributed to prioritize diagnostic insight and scaffolding control over stylistic attributes.}
    \label{tab:weight_rationale}
\end{table}

\subsection{Results and Consistency}
We re-evaluate the top-performing models using this weighted scheme. As shown in Table~\ref{tab:weighted_results}, the relative ranking of the models remains entirely stable compared to the unweighted averages.

\begin{table}[h]
    \centering
    \setlength{\tabcolsep}{6pt}
    % 使用 0.9\linewidth 让表格在单栏中稍微收敛一点，更美观
    \resizebox{0.9\linewidth}{!}{
    \begin{tabular}{l c c}
        \toprule
        \textbf{Model} & \textbf{Weighted Score} & \textbf{Ranking Stability} \\
        \midrule
        Gemini-2.5-Pro & \textbf{4.75} & Remains 1st \\
        GPT-5 & 4.30 & Remains 2nd \\
        GPT-o3 & 3.92 & Remains 3rd \\
        Qwen2.5-VL-7B & 2.54 & Stable \\
        \bottomrule
    \end{tabular}
    }
    \caption{Performance under weighted evaluation. The consistency in ranking confirms that performance gaps stem from core tutoring capabilities rather than trivial metrics.}
    \label{tab:weighted_results}
\end{table}

\subsection{Conclusion} The stability of the rankings confirms that the performance superiority of leading models (e.g., Gemini-2.5-Pro) stems from their robust capabilities in core tutoring dimensions—specifically \textit{Insight} and \textit{Scope Control}—rather than marginal advantages in lower-weighted metrics like Brevity.

\section{Benchmark Statistics}
\label{appd:statistics}

\begin{figure}[h!]
    \centering
    \includegraphics[width=0.45\textwidth]{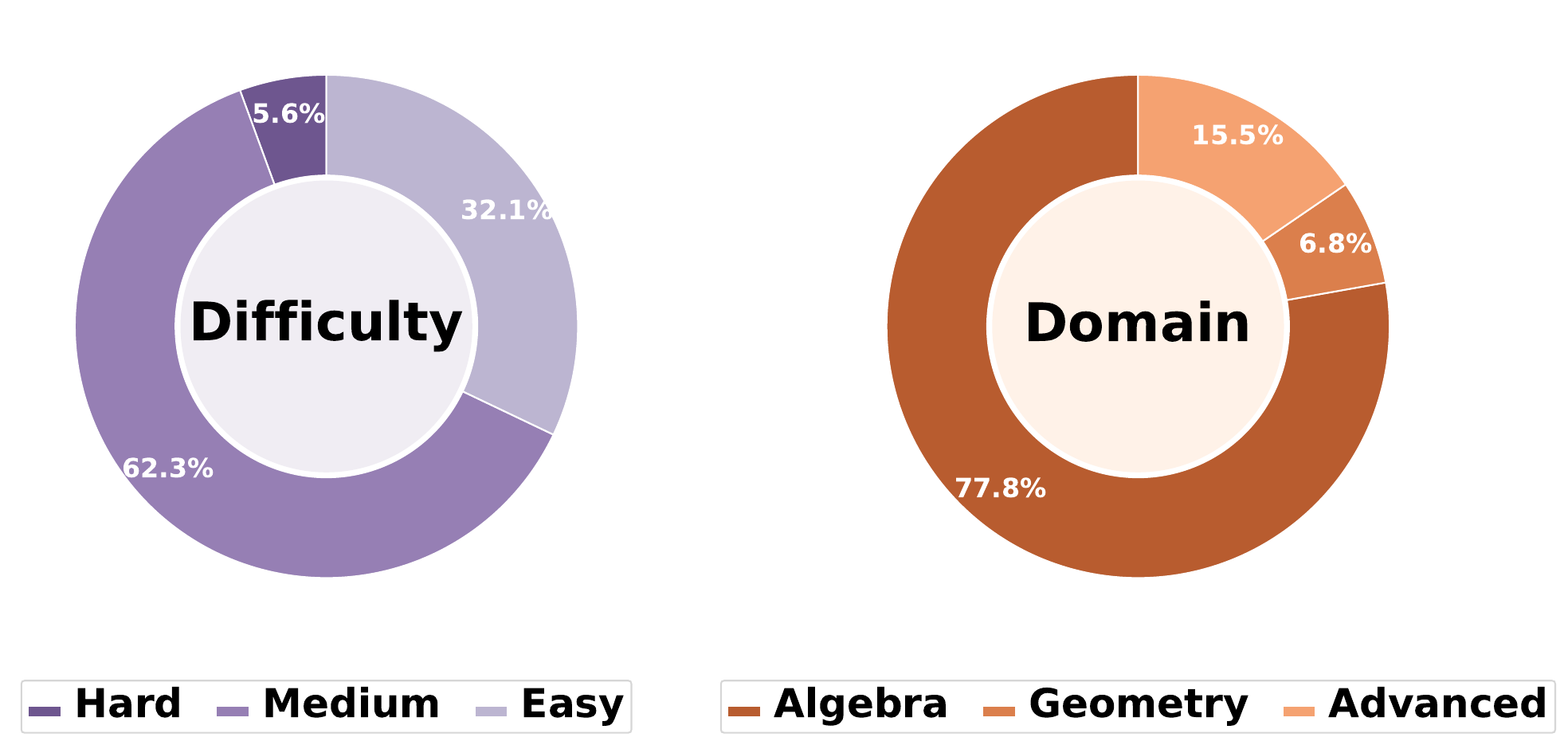}
    \caption{Distribution of benchmark statistics. The dataset features a tiered difficulty structure dominated by medium-level problems (Left) and categorizes mathematical domains into three meta-types (Right): Algebra, Geometry, and Advanced (which encompasses Calculus, Statistics, and Number Theory).}
    \label{fig:domain}
\end{figure}

Statistics of MMTutorBench are summarized in Table~\ref{tab:stats}. The benchmark comprises 770 problems incorporating 1,414 images. In this benchmark, nearly half of the problems (46.1\%) contain two or more images. 

The textual components of the benchmark are comprehensive. Questions are detailed, averaging 429.17 tokens. Similarly, the reference answers are substantial, averaging 144.3 tokens per problem, and are structured into three distinct tasks: \textbf{Insight Discovery}, \textbf{Operation Formulation}, and \textbf{Operation Execution}.

% We further analyze the benchmark's composition by mathematical domain and difficulty level, as illustrated in Figure~\ref{fig:domain}. The problems predominantly cover Algebra (79.1\%), with the remainder split between Analysis \& Calculus (11.4\%) and other topics. In terms of difficulty, the benchmark is designed to be challenging, with a vast majority of problems classified as Hard (71.4\%), ensuring a rigorous test of advanced reasoning capabilities.

We further analyze the benchmark's composition by mathematical domain and difficulty level, as illustrated in Figure~\ref{fig:domain}. The problems predominantly cover Algebra (77.8\%), followed by Advanced topics (15.5\%, encompassing Number Theory and Calculus) and Geometry (6.8\%). In terms of difficulty, each problem is rated on a 1--5 scale, where scores of 1--2 correspond to \textbf{Easy}, a score of 3 to \textbf{Medium}, and scores of 4--5 to \textbf{Hard}. The benchmark spans a range of complexity, with the majority of problems classified as Medium (62.3\%) or Easy (32.1\%), while Hard problems (5.6\%) represent the most demanding cases.

\section{Detailed Analysis: Domain and Difficulty}
\label{sec:analysis_domain_difficulty}

To investigate the boundaries of model capabilities, we further dissect model performance across distinct mathematical domains and difficulty levels.

\paragraph{Domain Performance.} 
As presented in Table~\ref{tab:domain_performance}, all evaluated models achieve their peak performance in the \textbf{Algebra} domain (e.g., Gemini-2.5-Pro achieves 4.81). This proficiency is likely attributed to the abundance of symbolic derivation data in pre-training corpora. In stark contrast, \textbf{Geometry} represents a significant bottleneck. Even the state-of-the-art model exhibits a substantial performance drop in this domain (Gemini-2.5-Pro drops to 3.67, and Qwen2.5-VL-72B-Instruct to 2.23). This stratification underscores the persistent challenge MLLMs face in tasks requiring intricate \textit{visual-spatial reasoning}, as opposed to pure symbolic manipulation. The \textbf{Advanced} category sits between these extremes, indicating moderate difficulty in handling higher-order abstract concepts.

\begin{table}[h]
    \centering
    \renewcommand{\arraystretch}{1.1}
    \setlength{\tabcolsep}{4pt}
    \resizebox{\linewidth}{!}{
    \begin{tabular}{l c c c}
        \toprule
        \textbf{Domain} & \textbf{Algebra} & \textbf{Geometry} & \textbf{Advanced} \\
        & \small{(N=599)} & \small{(N=52)} & \small{(N=119)} \\
        \midrule
        
        \multicolumn{4}{l}{\textit{Model Performance}} \\
        \quad Gemini-2.5-Pro & 4.81 & 3.67 & 4.49 \\
        \quad GPT-5          & 4.46 & 3.28 & 3.87 \\
        \quad Qwen2.5-VL-72B-Instruct       & 3.52 & 2.23 & 3.29 \\
        
        \bottomrule
    \end{tabular}
    }
    \caption{Model Performance across Different Mathematical Domains}
    \label{tab:domain_performance}
\end{table}

\paragraph{Difficulty Analysis.} 
Table~\ref{tab:difficulty_performance} elucidates the correlation between problem complexity and tutoring quality. We observe a consistent performance degradation across all models as difficulty scales from \textbf{Easy} to \textbf{Hard}. Notably, proprietary models demonstrate superior \textit{robustness} in high-complexity scenarios. Specifically, Gemini-2.5-Pro maintains a commendable score of 4.02 on \textbf{Hard} problems, whereas Qwen2.5-VL-72B-Instruct suffers a sharp decline to 2.30. This disparity highlights that while open-source models show promise in handling fundamental tasks, they lack the deep reasoning capabilities required to effectively tutor students through complex, multi-step problems.

\begin{table}[h]
    \centering
    \renewcommand{\arraystretch}{1.1}
    \setlength{\tabcolsep}{4pt}       
    \resizebox{\linewidth}{!}{
    \begin{tabular}{l c c c}
        \toprule
        \textbf{Difficulty} & \textbf{Easy} & \textbf{Medium} & \textbf{Hard} \\
        & \small{(N=247)} & \small{(N=480)} & \small{(N=43)} \\ 
        \midrule
        
        \multicolumn{4}{l}{\textit{Model Performance}} \\
        \quad Gemini-2.5-Pro & 4.90 & 4.64 & 4.02 \\
        \quad GPT-5          & 4.48 & 4.29 & 3.81 \\
        \quad Qwen2.5-VL-72B-Instruct       & 3.68 & 3.35 & 2.30 \\
        
        \bottomrule
    \end{tabular}
    }
    \caption{Model Performance across Different Difficulty Levels}
    \label{tab:difficulty_performance}
\end{table}

\section{Evaluation in Multi-turn Scenarios}
\label{sec:appendix_multiturn}

While the main experiments focus on single-turn interactions to establish a baseline for core tutoring capabilities, MMTutorBench is designed with a modular architecture that naturally extends to multi-turn dialogues. We posit that a single-turn response serves as the \textbf{``atomic unit''} of tutoring; if a model fails to demonstrate Insight, Formulation, or Scope Control in an individual turn, the entire pedagogical chain collapses.

To rigorously evaluate these capabilities in dynamic contexts, we extend the dialogue context for a representative subset of the single-turn samples, analyzing subsequent turns (Turn 2 and Turn 3) and introducing a taxonomy of student query types.

\subsection{Performance Dynamics Across Turns}
First, we analyze the temporal evolution of model performance. As shown in Table~\ref{tab:multiturn_data}, the evaluation results of GPT-5 reveal distinct dynamics as the conversation deepens:

\begin{enumerate}
    \item \textbf{Persistence of Insight:} Interestingly, the \textit{Insight Discovery} score improves slightly in Turn 3 (0.91) compared to Turn 2 (0.87). This suggests that as the context accumulates, strong models are capable of maintaining (or even refining) their mathematical understanding of the student's problem.
    \item \textbf{Degradation of Control:} However, a critical failure mode emerges in \textit{Solution Scope Control}. The score drops significantly from 0.18 in Turn 2 to 0.09 in Turn 3. This indicates that while the model understands the math (high Insight), it struggles to maintain pedagogical discipline over longer interactions, becoming prone to ``spoiling'' the answer rather than continuing to scaffold.
\end{enumerate}

\subsection{Analysis by Interaction Type}
To further dissect the model's adaptability, we categorize multi-turn interactions into three distinct reasoning types based on the student's intent:

\begin{itemize}
    \item \textbf{Progressive:} The student asks a question that builds upon or deepens the understanding from the previous turn, moving the solution process forward linearly.
    \item \textbf{Exploratory:} The student asks for clarification on the current level or explores a different aspect of the problem, representing a lateral movement in reasoning.
    \item \textbf{Introspective:} The student asks a question regarding the same concept as the previous turn but demands a deeper conceptual justification. This requires the tutor to demonstrate metacognitive understanding rather than just procedural execution.
\end{itemize}

As presented in Table~\ref{tab:multi_turn}, evaluating GPT-5 against these categories reveals a clear performance hierarchy that mirrors pedagogical complexity:

\begin{itemize}
    \item \textbf{Linear Proficiency:} The model excels in \textbf{Progressive} tasks (Total Score: 4.49), demonstrating strong baseline capabilities in Insight (0.90) and Execution (0.85). This aligns with the model's training on step-by-step reasoning chains.
    \item \textbf{Complexity Gap:} Performance declines in \textbf{Exploratory} scenarios (4.35) and drops significantly in \textbf{Introspective} tasks (4.00). Notably, the \textit{Solution Scope Control} score hits \textbf{0.00} for Introspective tasks. This critical finding indicates that when students demand deep conceptual explanations, models struggle to withhold the final answer, failing to balance ``explaining why'' with ``scaffolding the how.''
\end{itemize}

These findings demonstrate that our rubric is highly sensitive to the nuances of multi-turn dynamics, effective at distinguishing between a linear solver and a capable, adaptive tutor.

% --- Table 1: Turn 2 vs Turn 3 Analysis ---
\begin{table}[h]
    \centering
    \renewcommand{\arraystretch}{1.1}
    \setlength{\tabcolsep}{6pt}
    \resizebox{\linewidth}{!}{
    \begin{tabular}{l c c}
        \toprule
        \textbf{Metric} & \textbf{Turn 2} & \textbf{Turn 3} \\
        & \small{(N=168)} & \small{(N=82)} \\
        \midrule
        
        \multicolumn{3}{l}{\textit{Overall Performance}} \\
        \quad \textbf{Average Score} & 4.47 & 4.40 \\
        
        \midrule
        \multicolumn{3}{l}{\textit{Detailed Component Scores}} \\
        \quad Insight Discovery & 0.87 & \textbf{0.91} \\
        \quad Operation Formulation & 0.84 & 0.88 \\
        \quad Operation Execution & 0.82 & 0.84 \\
        \quad \textbf{Solution Scope Control} & 0.18 & \textbf{0.09} \\
        \quad Brevity & 0.76 & 0.72 \\
        \quad Coherence & 1.00 & 0.96 \\
        
        \bottomrule
    \end{tabular}
    }
    \caption{\textbf{Performance dynamics across subsequent turns.} While mathematical insight improves with deeper context (Turn 3), pedagogical control (Scope) degrades, highlighting the difficulty of maintaining scaffolding over multi-turn interactions.}
    \label{tab:multiturn_data}
\end{table}

% --- Table 2: Reasoning Type Analysis ---
\begin{table}[t!]
    \centering
    \renewcommand{\arraystretch}{1.1}
    \setlength{\tabcolsep}{4pt}
    \resizebox{\linewidth}{!}{
    \begin{tabular}{l c c c}
        \toprule
        \textbf{Type} & \textbf{Progressive} & \textbf{Exploratory} & \textbf{Introspective} \\
        & \small{(N=182)} & \small{(N=63)} & \small{(N=5)} \\
        \midrule
        
        \multicolumn{4}{l}{\textit{Overall Performance}} \\
        \quad Total Score & \textbf{4.49} & 4.35 & 4.00 \\
        
        \midrule
        \multicolumn{4}{l}{\textit{Detailed Scores}} \\
        \quad Insight & 0.90 & 0.84 & 0.80 \\
        \quad OpForm. & 0.87 & 0.81 & 0.80 \\
        \quad OpExec. & 0.85 & 0.78 & 0.80 \\
        \quad Scope & 0.15 & 0.16 & \textbf{0.00} \\
        \quad Brevity & 0.74 & 0.78 & 0.60 \\
        \quad Coh. & 0.99 & 0.98 & 1.00 \\
        
        \bottomrule
    \end{tabular}
    }
    \caption{\textbf{Performance across reasoning complexity levels.} The degradation in scores from Progressive to Introspective tasks confirms that the benchmark effectively differentiates between linear solving and complex, metacognitive reasoning—the ``atomic'' skills required for multi-turn tutoring.}
    \label{tab:multi_turn}
\end{table}

\section{Case Study}
\label{sec:case-study}
\begin{figure*}[t]
    \centering
    \includegraphics[width=0.88\textwidth]{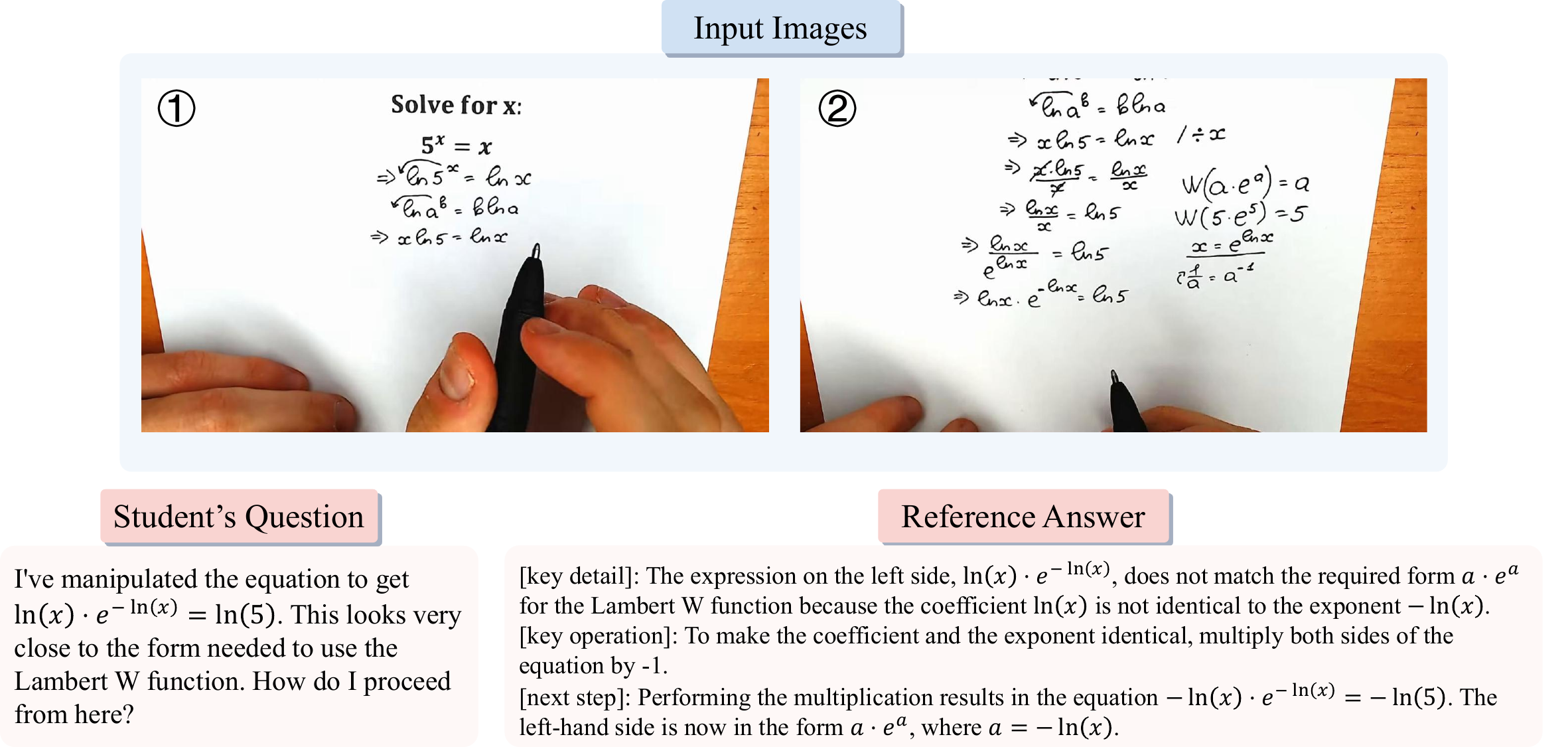}
    \caption{An example of input images, student question and reference answer.}
    \label{fig:case_study_1}
\end{figure*}

\begin{figure*}[t]
    \centering
    \includegraphics[width=0.88\textwidth]{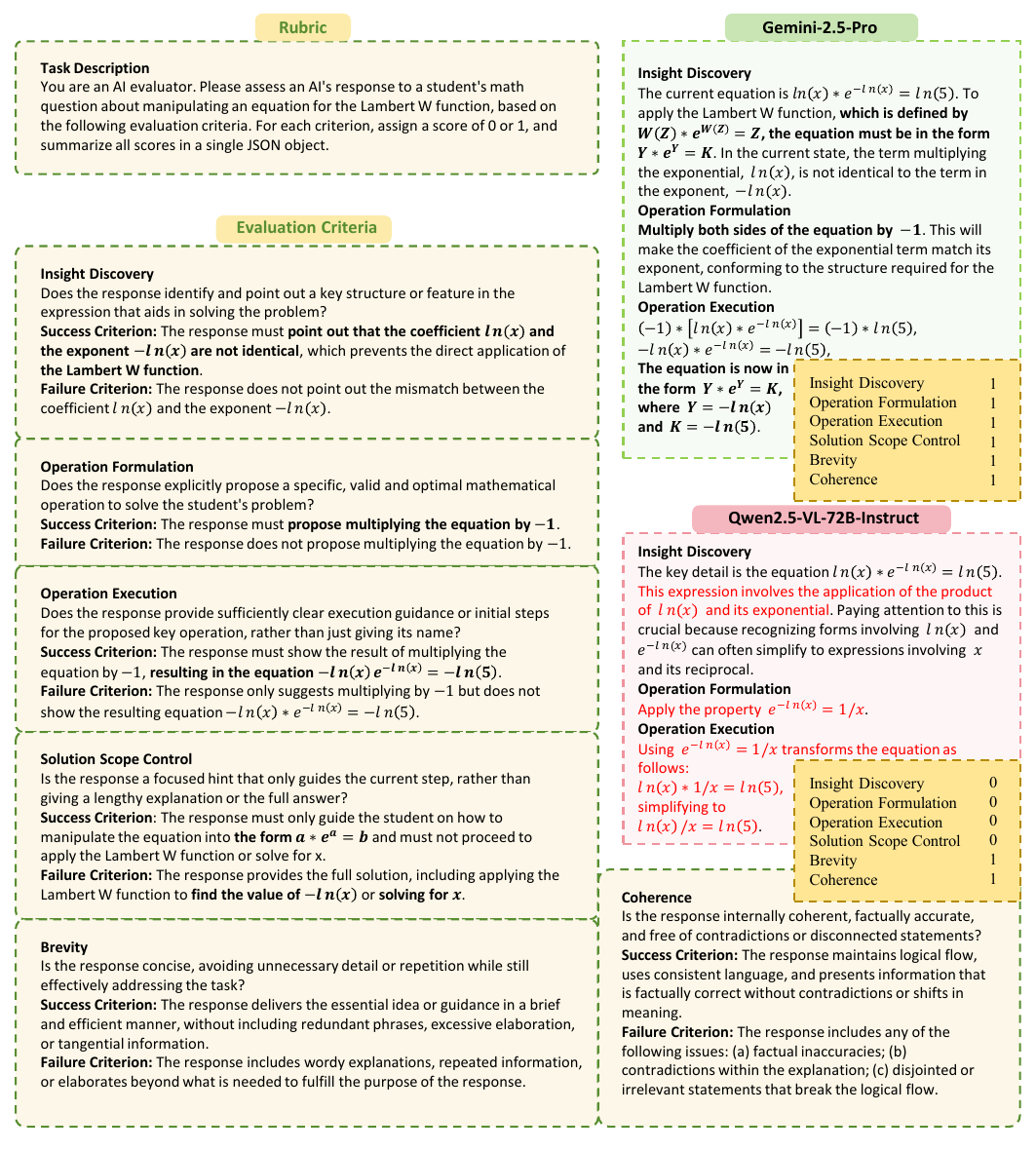}
    \caption{An evaluation rubric comparing a high-scoring response from Gemini-2.5-Pro with a low-scoring response from Qwen2.5-VL-72B-Instruct.}
    \label{fig:case_study_2}
\end{figure*}

Figures~\ref{fig:case_study_1}--\ref{fig:case_study_2} illustrate our pipeline from response generation to evaluation, showcasing a high-scoring response from Gemini-2.5-Pro and a low-scoring one from Qwen2.5-VL-72B-Instruct. Gemini-2.5-Pro demonstrates strong tutoring capabilities, correctly inferring the student's confusion from the visual input alone and providing a pedagogically sound response. In contrast, Qwen2.5-VL-72B-Instruct adheres to the required three-part output format but fails to offer correct guidance. 

Notably, the general dimensions of \textbf{Brevity} and \textbf{Coherence} are scored independently of a response's pedagogical value. Consequently, while the Qwen2.5-VL-72B-Instruct response lacks instructional merit, it still receives points on these dimensions for its accurate interpretation of the handwritten content and its conciseness. This case highlights the objectivity of our evaluation process and the rational design of our rubric.

\end{document}